\documentclass{IEEEtran}
\usepackage{wrapfig}
\usepackage{cite}
\usepackage{amsmath,amssymb,amsfonts}
\usepackage{algorithm,algorithmic}
\usepackage{siunitx}
\usepackage{graphicx}
\usepackage{textcomp}
\usepackage{xcolor}
\usepackage{placeins}
\usepackage{hyperref}
\def\BibTeX{{\rm B\kern-.05em{\sc i\kern-.025em b}\kern-.08em
    T\kern-.1667em\lower.7ex\hbox{E}\kern-.125em}}
\usepackage[normalem]{ulem}
\usepackage{xcolor}
\usepackage{ifthen}

\newcommand{\Highlit}[1]{\textcolor{red}{#1}}

\begin{document}
\title{Deep Autoencoders with Value-at-Risk Thresholding for Unsupervised Anomaly Detection}
%\title{Some Practical Aspects of Background Modelling}
%\title{On comparison of background modelling algorithms}
\author{
\IEEEauthorblockN{Albert Akhriev}\IEEEauthorblockA{IBM Research -- Ireland}
\and
\IEEEauthorblockN{Jakub Marecek}\IEEEauthorblockA{IBM Research -- Ireland}
}
\author{Albert Akhriev and Jakub Marecek}%˜\IEEEmembership{ Member, IEEE,}%
%}
\maketitle
%-------------------------------------------------------------------------

\begin{abstract}
Many real-world monitoring and surveillance applications require non-trivial anomaly detection to be run in the streaming model. We consider an incremental-learning approach, wherein a deep-autoencoding (DAE) model of what is normal is trained and used to detect anomalies at the same time. 
In the detection of anomalies, we utilise a novel thresholding mechanism, based on value at risk (VaR). 
We compare the resulting convolutional neural network (CNN)  against a number of subspace methods, and present results on changedetection.net.
\end{abstract}

% For peer review papers, you can put extra information on the cover page as needed:
% \fi
\IEEEpeerreviewmaketitle

%-------------------------------------------------------------------------
\section{Introduction}
\label{sec:intro}
%-------------------------------------------------------------------------

Consider the problem, where starting from high-dimensional streamed data, one should like to distinguish between ``normal'' input of time-varying nature (also known as background) and  ``anomalies'' (events, or foreground). 
This problem is known as anomaly or event or outlier detection in Data Engineering \cite{5767923,7930039} and as background subtraction in Computer Vision \cite{Bouwmans19}. 

Early methods studied the data (block-)coordinate-wise. For example for image data, these methods worked pixel-by-pixel  \cite{Stauffer1999,Zivkovic06,Barnich09,Bergevin15}, where each pixel is a block of three coordinates.
Subsequently, the view of anomaly detection as a low-rank matrix-completion problem has become popular. 
A typical approach stacks a number of recent flattened observations, e.g. video frames, into rows of a data matrix, which is then approximated via a low-rank matrix. 
The low-rank model corresponds to the background,
and anomalies are outside of the low-rank subspace  
\cite{Lin11,Balzano13,Rodriguez13,He11,Oreifej13}. 
In Signal Processing \cite{Vaswani18,Chen18,Lerman18}, there is much related work on robust principal component analysis (RPCA) and subspace tracking.
Extensive literature surveys can be found in \cite{Ma18,Bouwmans18,Balzano18}. 

More recently, deep-learning techniques have been developed, based on the matrix-completion view of anomaly detection \cite{7930039,7930041}. Notably, autoencoder architecture have proven successful, in practice. A multi-scale framework, proposed in \cite{Lim18}, encodes the input images by means of pre-trained VGG-16 followed by a sub-net that pools features at multiple scales before feeding them into decoder. Authors claim robustness against camera jitter and shadows despite a very limited number of labelled images used for training. In \cite{Babaee17} authors extended ideas of \cite{Stauffer1999,Zivkovic06} by training a CNN on image patches. Subsequently, the trained CNN was applied to assemble foreground mask from patches on previously unseen images. Likewise a patch-based CNN in \cite{Braham16} learns to output foreground probability on a small number of labelled training examples. Double-autoencoder network, introduced in \cite{Xu14}, reconstructs background in two stages. It requires some initial training but afterwards can proceed in online fashion. 
There are also sparse \cite{lee2007efficient}, robust \cite{zhou2017anomaly} variants and variants combining auto-encoders with Gaussian mixture models \cite{zong2018deep}. 
For a broader view of recent advances in deep architectures for background modelling, reader is advised to consult the comprehensive surveys \cite{Bouwmans18nn,Bouwmans19}.

A key challenge across subspace and deep-learning approaches is the amount of supervision and tuning. 
Interestingly, many approaches therein rely on very simple thresholding mechanisms that require extensive tuning and output quite noisy foreground masks. 
In particular, across both traditional methods \cite{Stauffer1999,Zivkovic06},
methods based on matrix-completion \cite{Yong18},
and autoencoders \cite{zong2018deep}, 
the use of Gaussian mixture models (GMM) is the state of the art.
While there are plausible alternatives \cite[e.g.]{8635825}, 
the typical use of GMM involves the use of expectation-maximization (EM) heuristics, 
which suffer from a host of issues, including the sensitivity to noise and sensitivity to balance
in the mixing coefficients, as well as getting stuck in arbitrarily poor local optima.
One would hence like to obtain an unsupervised approach, without the GMM assumptions.

Our main contribution is a technique for unsupervised use of deep autoencoders:
\begin{itemize}
    \item the use of a thresholding mechanism based on value at risk (VaR), which can be computed exactly in time required to sort the incoming data.
    \item a novel weighing (pre-processing) of the input to the autoencoder. 
    \item a numerical study of deep-autoencoders and matrix-completion methods with a variety of thresholding methods on changedetection.net.
\end{itemize}
The use of VaR-based thresholding makes it possible to adapt deep-learning approaches into unsupervised methods without data-dependent tuning.
%will behave in real applications with a variety of adversarial factors.

The paper is organised as follows: in Section~\ref{sec:algo}, we describe a variant of background subtraction algorithm based on a convolutional autoencoder. In Section~\ref{sec:exp}, it is compared against selected low-rank approximation methods. Both approaches can be seen as non-linear subspace tracking \cite{hinton2006reducing}, which motivated our choice of comparison candidates. Section~\ref{sec:conclude} summarises our vision of future efforts.

%-------------------------------------------------------------------------
\section{The Unsupervised Convolutional Autoencoder}
\label{sec:algo}
%-------------------------------------------------------------------------

Since having labelled data is unaffordable in many real-world applications, an unsupervised approach to anomaly detection is desirable.
Our approach uses a convolutional neural networks (convolutional autoencoder) without an explicit training phase
 for anomaly detection in streamed data. 
In particular, it uses incremental training of a model of what is normal (background), without any supervised data, and concurrent use of the model, to estimate what is normal (background). 

A baseline deep convolutional autoencoder (DAE) could be seen as a generalisation of low-rank approximation methods.
Upon arrival of a new frame $I$ of a stream, we do exactly one forward/backward iteration in order to train the autoencoder (\textit{update phase}), and then draw an estimate of background model $B$ as an output of the autoencoder network (\textit{reconstruction phase}). $L_1$-norm loss function minimises the difference between $I$ and $B$, effectively ignoring the outliers (here, points of moving objects).

\begin{figure*}
\begin{center}
\includegraphics[width=0.45\textwidth, angle=90]{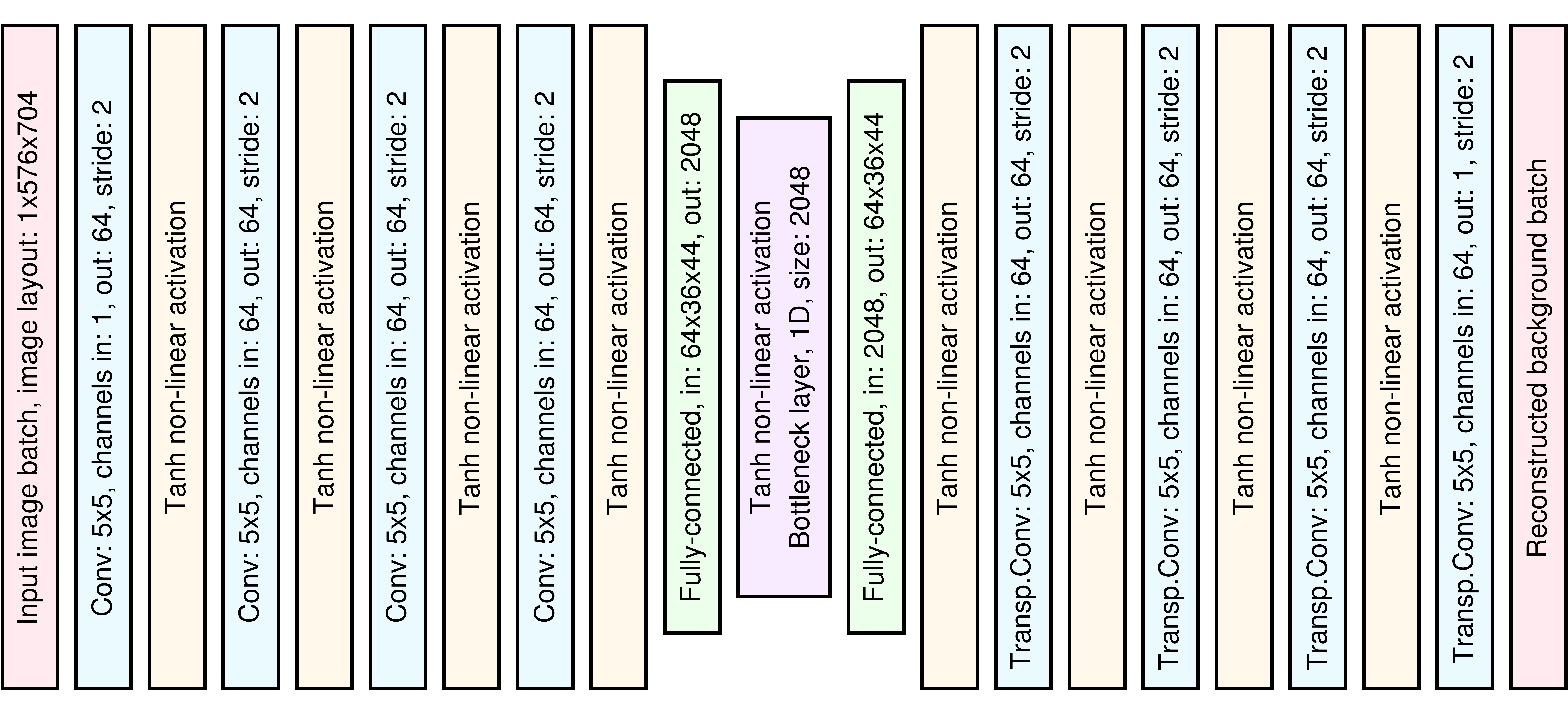}
\end{center}
\caption{The architecture of the convolutional autoencoder. Note that after the second fully-connected layer, we reshape the intermediate tensor to ${64{\times}36{\times}44}$ for subsequent decoding, where the number of channels (64) comes first in PyTorch convention. Also, parameter ``padding=2'' is specified in all convolutional layers and additionally ``output\_padding=1'' in transposed convolutional layers (Transp.Conv) respectively.}
\label{fig:network}
\end{figure*}

Figure~\ref{fig:network} depicts the architecture of our autoencoder. Except the first and the last layer, all other convolution layers have 64 filters of size $5{\times}5$ (input and output number of channels equals to 64), interleaved with non-linear activations. Experimentally, we found that hyperbolic tangent function (tanh) works better than logistic function or rectified linear unit.

In a pre-processing step, the input is flattened, block-coordinate to a single coordinate  (e.g., red-green-blue to grayscale), before being fed into the autoencoder. (Note that at night time, there are no colours in video data anyway.) Each frame is then reshaped into a ``standard'' layout ($1{\times}576{\times}704$) and the values are normalised to $[-0.5 \ldots 0.5]$ range. Reconstructed background and foreground mask undergo the inverse transformation. The bottleneck layer is represented by 1D tensor of size $2048$, surrounded by fully-connected layers. Encoder's layers are shrunk by a factor of $2$ (stride $2$), as the data propagate from input to output, and decoder's (transposed) layers are expanded by the same factor of $2$ (stride $2$), respectively. Here we try to balance between network depth (that increases computational burden) and background reconstruction quality.

In early experiments, we used single image as an input and output respectively. This approached demonstrated solid results compared to other methods. Nonetheless, dynamic background case can be handled  better, if we admit multiple images in the process of outlier thresholding. We have at least two options here. First, accumulate a short history of recent frames (50 in our experiments, or 2 seconds of video). In each iteration, pick up a subset of 10 images uniformly spread over this short history, and stack them into $10{\times}H{\times}W$ tensor, where $W$, $H$ stand for width and height respectively. This tensor is fed into an autoencoder as a multi-channel image $I(c,i)$  with ``colour channels'' $c = 1 \ldots 10$ and points $i = 1 \ldots (H \times W)$. The reconstructed background $B(c,i)$ has the same layout. The goal is to make training procedure less prone to overfitting.
% $I(c,x,y)$

Multi-channel input image increases the number of weights in the first and the last convolution layers making forward/backward passed more compute-intensive. The second option, also adopted in this study, presents a trade-off between CPU/GPU load and foreground detection quality. Namely, we use approach similar to described above except a multi-channel input is replaced by a mini-batch of 10 input/output images. This reduces the number of weights in convolutional layers but, in theory, also lessens the flexibility of autoencoder network. In practice, we found no difference and both options produce very similar results up to minor variations attributed to randomization in Xavier's initialization of network weights.

Algorithm~\ref{alg:autoencoder} summarizes the main steps of proposed approach 1) transform a new frame into a ``standard'' form; 2) update a time window of a history of 50 recent frames; 3) compute weights using the optic-flow algorithm (\ref{eq:weight}) and plug them into the loss function (\ref{eq:loss}); 4) make one forward and one backward step in training the autoencoder (cf. Figure~\ref{fig:network}); 5) reconstruct background model $B$, and compute the residuals; 6) estimate the optimum threshold using Value at Risk, and apply it to the residuals; 7) output a binary mask outlier/background reshaped back to the original size.

\begin{algorithm}[!htb]
\caption{A single step of the incremental autoencoder training and anomaly detection}
\label{alg:autoencoder}
\algsetup{indent=2em}
\begin{algorithmic}[1]
\STATE{\textbf{Input}: Next element from a data stream, e.g., one frame of video data.}
\STATE{\textbf{Output}: a binary mask suggesting what is an anomaly.}
\STATE{Reshape the data to suit the convolutional network and normalise the values, e.g., 
to an image in  $1{\times}576{\times}704$ resolution,
linearly transformed to $[-0.5 \ldots 0.5]$ range.}
\STATE{Update a time window considered, e.g., a history of 50 recent images.}
\STATE{Compute weights (\ref{eq:weight}) using the optic flow and plug them into the loss function (\ref{eq:loss}).}
\STATE{Make one forward and one backward step in training the autoencoder.}
\STATE{Reconstruct a model of what is normal  (background) $B$, and compute residuals $r_i$.}% (\ref{eq:res}).}
\STATE{\textbf{return} {\tt threshold}($r_i$) of Algorithm \ref{alg:thresholding}, which estimates the optimum threshold, and applies it to the  residuals.}
\end{algorithmic}
\end{algorithm}

\subsection{DIS-Based Weights} 

As pointed out in \cite{Xu14}, 
the robustness of the loss function $L$ can be improved by introducing point-wise weights:
\begin{equation}
L = \sum\nolimits_{c,i} w_i \left|I(c, i) - B(c, i)\right|,
\label{eq:loss}
\end{equation}
where index $i$ runs through all the pixels, and $w_i$ is close to 0 for a moving point, and close to 1 otherwise. 
Because filters in CNN are shared by all the points, few remaining outliers would not harm the training process and will be suppressed by $L_1$-norm loss function. One possible option for weight computation is to decrese the weights using the DIS algorithm \cite{Kroeger16}:
\begin{equation}
w_i = \exp{\left(-v_i^2/\left(2\cdot\mbox{median\_over\_image}\left(v^2\right)\right)\right)},
\label{eq:weight}
\end{equation}
where $v_i$ is a velocity at point obtained by optic-flow computation from a pair of consecutive input frames $I^{(t-1)}$ and $I^{(t)}$, at times $t-1, t$. The scaling factor inside the exponent is computed as a median squared velocity over all points. In this way, the moving points are effectively down-weighed during the training phase. 
%Since ``outliers'' are rare by assumption, outliers appear only temporarily and pixels masked out on a subset of frames become observable again as soon as a moving object passed by. 
Alternatively, the masks output by fast and reliable algorithms introduced in \cite{Barnich09,Bergevin15} could be used as the weights. Note that it would be sufficient to mask out the majority of foreground points. Because filters in a convolutional network are shared by all the points, few remaining outliers would not be harmful for the training process and will be suppressed by $L_1$-norm loss function.

\section{The VaR-Based Thresholding} 
\label{app:threshold}

A key step in the use of low-rank or autoencoding approaches is thresholding. 
Therein, one considers the so-called \textit{residual map}, which is an array of the same dimensions as the input $I$ and background $B$.
In this study, the residual $r_i$ at $i$th point is defined by:
\begin{equation}
r_i = \min_{k \in {\cal N}_i,\,\,c = 1...10}\Big|I(1,i) - B(c, i)\Big|,
\label{eq:residvalue}
\end{equation}
wherein ${\cal N}_i$ are the points in $3{\times}3$ vicinity of (central) point $i$, $c$ ranges over 10 recently reconstructed backgrounds,
and a variety of norms (distance functions) can be used, outside of the absolute value of (\ref{eq:residvalue}). 
Based on the residual map, the thresholding produces a binary-per-block-coordinate array $r_i > t$, which suggest what are the anomalies (moving objects) and what constitutes the normal background process.

We should like to stress that 
thresholding is a vast area by itself, even when restricted to anomaly detection \cite{Sezgin04}. Although locally adapted threshold, e.g. \cite{Bradley07,Tao08}, may work best, it is quite common to choose a single, even hard-coded, threshold for each frame. We follow the same practice by choosing automatically estimated, global threshold for every block-coordinate in residual map.
Among the techniques for automatic threshold selection, we found that classical methods and their variations such like Otsu's one \cite{Otsu79}, where threshold is determined by minimizing intra-class intensity variance, or maximum entropy method in \cite{Kapur85} do not produce convincing foreground/background segmentation. The comprehensive surveys \cite{Rosin98,Sezgin04,Chang06} give a good insight into the related methods.

Apparently, a threshold estimator should take some spatial information into consideration in order to make an ``optimal'' decision. At a high level, we seek a threshold for the \textit{highest sensitivity}, when isolated noisy points ``just'' show up. To this end, we need to define isolation and sensitivity.

\begin{figure}[H]
\begin{center}
\includegraphics[width=0.48\textwidth]{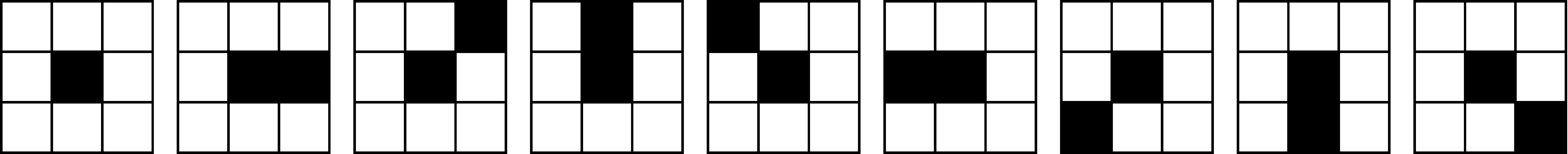}
\caption{The configurations of the $3 \times 3$ contiguous patches, whose fraction within all the $3 \times 3$ contiguous patches is sought.}
\label{fig:configs}
\end{center}
\end{figure}

In defining isolation, we have adopted a simple, yet efficient, mechanism  first proposed by Malistov  \cite{Malistov14}. He estimated the probability of formation of a false ``object'' when $4$ or more adjacent points exceeded a threshold. From that consideration, he deduced an optimum threshold value. We extended the idea by testing spatial neighborhood of a pixel and examining $3{\times}3$ contiguous patches that have $1$ or $2$ pixels, including the center, marked as an anomaly. This is illustrated in Figure~\ref{fig:configs}. 

For a given threshold, one analyses $3 \times 3$ patches centered at each point in the residual map. There are several cases how residual value $r_i$ at the central point relates to its neighbours. Let us consider one example. Suppose that the central value $r_1$ is the largest one and we pick up the second $r_2$ and the third $r_3$ largest ones from the $3 \times 3$ patch, $r_3 \le r_2 \le r_1$. Suppose that all the values are integral, as usual in computer-vision applications. If a threshold happens to lie in the interval $[r_3+1 \ldots r_1]$, then one of the patterns depicted on Figure~\ref{fig:configs} will show up after thresholding. As such, this particular central point ``votes'' for the range $[r_3+1 \ldots r_1]$.

\begin{figure}[tb]
\begin{center}
\includegraphics[width=0.5\textwidth]{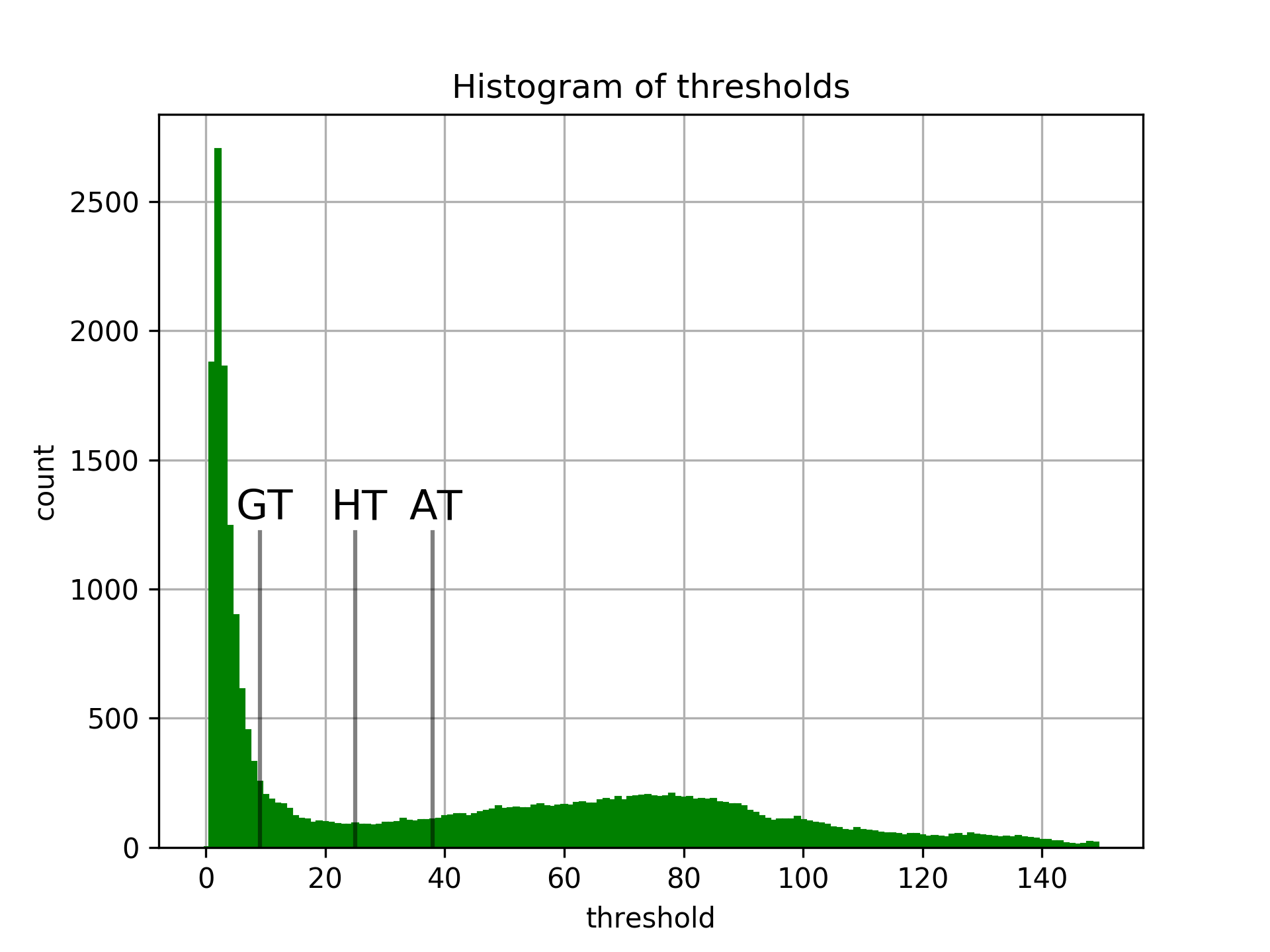}
\end{center}
\caption{An example of the histogram of thresholds. The histogram was truncated from the original $255$ size to allow for some clarity of presentation. Three threshold values read as follows: GT -- ``ground-truth'' threshold obtained by minimizing a mismatch between binarized image of residuals and the ground-truth foreground mask provided in cdnet~2014; HT -- ``hard'' threshold, here 25; AT -- ``automatic'' threshold obtained by Algorithm~\ref{alg:thresholding}.}
\label{fig:histthr}
\end{figure}

\begin{figure}[tb]
\begin{center}
\includegraphics[width=0.5\textwidth]{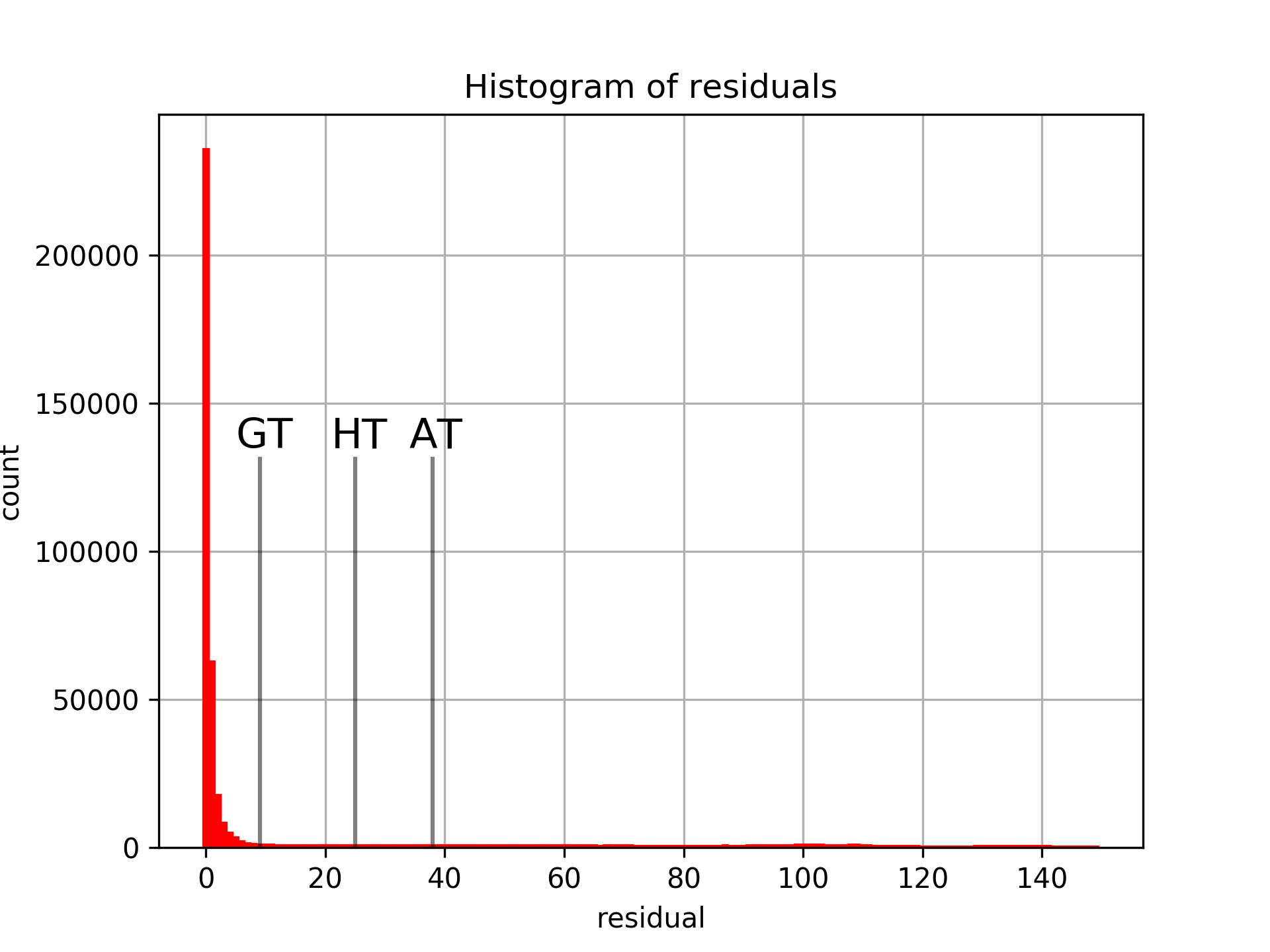}
\end{center}
\caption{An example of the histogram of residuals. The histogram was truncated from the original $255$ size to allow for some clarity of presentation. Three threshold values read as follows: GT -- ``ground-truth'' threshold obtained by minimizing a mismatch between binarized image of residuals and the ground-truth foreground mask provided in cdnet~2014; HT -- ``hard'' threshold, here 25; AT -- ``automatic'' threshold obtained by Algorithm~\ref{alg:thresholding}.}
\label{fig:histres}
\end{figure}

\begin{figure}[ht]
\begin{center}
\includegraphics[width=0.5\textwidth]{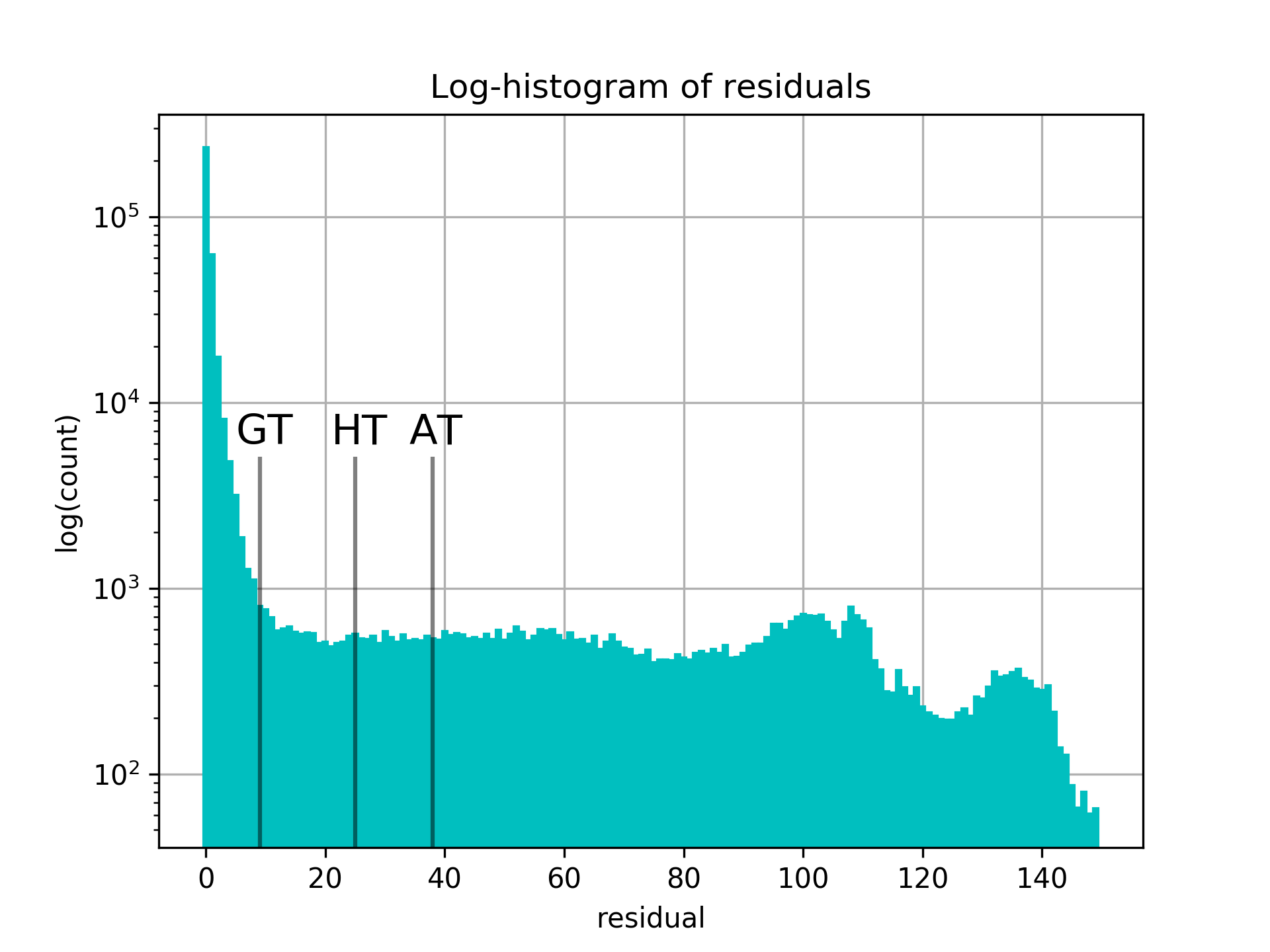}
\end{center}
\caption{An example of the histogram of residuals. The histogram was truncated from the original $255$ size to allow for some clarity of presentation. Also, the vertical axes was {\it log}-transformed making tail values better visible. Three threshold values read as follows: GT -- ``ground-truth'' threshold obtained by minimizing a mismatch between binarized image of residuals and the ground-truth foreground mask provided in cdnet~2014; HT -- ``hard'' threshold, here 25; AT -- ``automatic'' threshold obtained by Algorithm~\ref{alg:thresholding}.}
\label{fig:histreslog}
\end{figure}

One can view the ``votes'' as forming a probability mass function of a random variable supported on the range of possible values of the thresholds, e.g., [0, 255]. 
Figure~\ref{fig:histthr} suggests to see this as an \textit{histogram of thresholds}.
In particular for the center point of the previous paragraph, we would increment counters in the bins corresponding to threshold values $r_3+1$ to $r_1$ in the histogram. Repeating the process for all the points and all combinations of $3$ largest residuals $r_1$, $r_2$, $r_3$, we arrive at the information as to the number $H_T(t)$ of (central) points of one of the patterns of Figure~\ref{fig:configs} would be observed if the threshold $t$ were used. In other words, $H_T(t)$ gives a rate of appearance of patterns in Figure~\ref{fig:configs} as a function of selected threshold $t$. 
Figures~\ref{fig:histthr} to \ref{fig:histreslog} give an example of a histogram of thresholds, a histogram of residuals and a {\it log}-transformed histogram of residuals respectively. 

In defining sensitivity, consider a discrete random variable $X$, and risk measures thereof. 
While it is customary to analyze the histogram of residuals, e.g. \cite{Otsu79,Kapur85,Conaire06,Rosin98,Sezgin04,Fu-song14},
all our calculations are done over the histogram of thresholds introduced above.
As risk measures, we consider the value at risk (VaR):
\begin{align}
\textrm{VaR}_\alpha(X) := \min\{c : P(X \le c) \ge \alphaα \},
\end{align}
and the conditional value at risk (CVaR):
\begin{align}
\textrm{CVaR}_\alpha(X) := \mathbb{E}[X : X \ge \textrm{VaR}_{\alpha}(X)].
\end{align}
CVaR is also known as the
Average Value-at-Risk, Expected Shortfall, and Tail Conditional Expectation, in various communities within computational finance, where the random variable usually models the loss associated with an asset or a collection of assets.

\begin{figure*}
\begin{center}
\includegraphics[width=\textwidth]{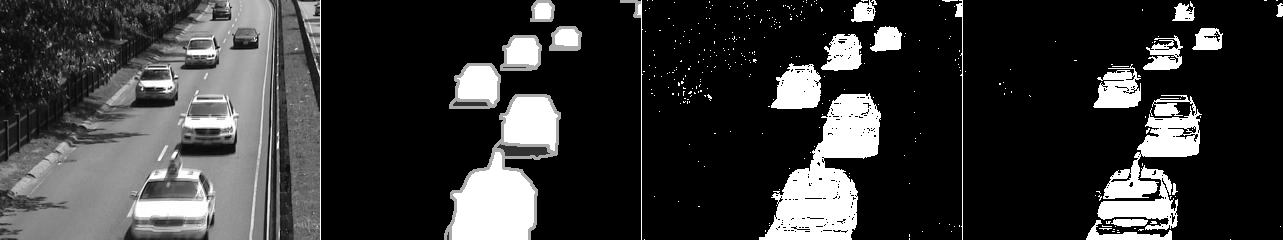}
\end{center}
\caption{Sample frame from the video-sequence \textit{highway} (left), corresponding ground-truth foreground mask provided by cdnet~2014 (middle-left), result of binarization of residual image by the ``ground-truth'' threshold obtained by minimizing a mismatch between binarized image and the ground-truth (middle-right), result of binarization of residual image by the ``automatic'' threshold obtained by Algorithm~\ref{alg:thresholding} (right).}
\label{fig:sampleframe}
\end{figure*}

For comparison purposes, we introduce two more types of thresholds in this study. The first one is called an ``ground-truth'' threshold (GT), which is obtained by minimizing a mismatch between thresholded residuals and the ground-truth foreground mask provided in cdnet~2014 dataset. A ``hard'' threshold (HT) is a hard-coded value of 25.
(Notice that a ``hard'' threshold is quite common in state-of-the-art-implementations, cf. \cite{lrslibrary2015}.)
For the sample frame in Figure~\ref{fig:sampleframe}, our automatic threshold is $\mbox{VaR}_{\alpha}^{(at)} = 38$ and 
the ground-truth threshold is $\mbox{VaR}_{\alpha}^{(gt)} = 9$. The corresponding parameters are $\alpha_{gt} = 0.835$ and $\alpha_{at} = 0.876$ respectively. Loosely speaking, this means the following: Had we selected the above VaRs as thresholds, a pixel with the residual exceeding the values of $83.5\%$ and $87.6\%$ of the smallest residuals should be considered as an outlier (a point of a moving object in this context).For the same parameters $\alpha$, the corresponding CVaR values are $\mbox{CVaR}_{\alpha}^{(gt)} = 70.1$ and $\mbox{CVaR}_{\alpha}^{(at)} = 85.8$ respectively. The big difference between VaR and CVaR values (for the same $\alpha$) can be explained by slow decay of the residual distribution, see Figure~\ref{fig:histreslog}, which contains not only the regular background samples, but also the outliers belonging to moving objects.

Figure~\ref{fig:sampleframe} shows the result of thresholding by either ``ground-truth'' or by our ``automatic'' threshold. Clearly, the ground-truth threshold produces a solid mask, but it captures the background noise as well. In some cases, it could be not trivial to eliminated that noise. On the other hand, the ``automatic'' threshold gives less amount of clutter in background, but more ``holes'' in moving objects. Note, we do not use any mask post-processing in this study. %Unfortunately, we can not avoid some of loss mask integrity while increasing a risk of acquiring a background noise. At least in the scenario of a single, global threshold applied to all the points. On the other hand, more realistic locally-adaptive thresholding could be an expensive solution.

%In addition, we set up a so called ``hard'' threshold --- empirically selected value $T_h = 25$. The latter approach is quite common and widely used in many implementations in \cite{lrslibrary2015}. It was found experimentally that binarization with the hard threshold $T_h = 25$ produces a reasonable foreground masks in the majority of situations, although in the future we would like to avoid any predefined values at all.

Algorithm~\ref{alg:thresholding} summarizes all the steps. It returns maximum of the four values: (1) hard threshold $T_h$; (2) value exceeding $2/3$ of the smallest residuals; (3) right-hand side margin of the smallest interval that contains $50\%$ of histogram area; (4) automatic VaR threshold as described above.

% While searching for the best threshold (from the right margin to the end of the histogram of thresholds), we take the next value $t$ and explore the range $[t\!-\!5\,\ldots\,t\!+\!5]$. If all the histogram values $H_T(x)$ are below $0.0025{\cdot}N$, where $N$ is the number of pixels and $x \in [t\!-\!5\,\ldots\,t\!+\!5]$, then we stop further searching and report the optimum threshold $t$. This window-based approach slightly improves the segmentation result. 

\begin{algorithm}[ht]
\caption{Computation of Value-at-Risk threshold}
\label{alg:thresholding}
\algsetup{indent=2em}
\begin{algorithmic}[1]
\STATE{\textbf{Input}: 
\begin{itemize}
    \item image of integral residuals $r_i \in [0, 255]$,
    \item rate of noisy patterns in Figure~\ref{fig:configs}, $R = 0.0025$,
    \item hard threshold $T_h = 25$.
\end{itemize}}
\STATE{\textbf{Output}:  threshold $t$.}
\STATE{Initialise threshold $t$ to a lower bound, which is just above the values of $2/3$ of the smallest residuals.}
\STATE{Build the histogram of thresholds $H_T(t)$ from $r_i$.}
\STATE{Compute the smallest interval that contains $50\%$ of the histogram area, and get the right margin of this interval $m_{right}$.}
\STATE{Adjust the initial threshold: $t \gets \max{\left(t, T_h, m_{right}\right)}$.}
\FOR{$t \le 255$}
    \STATE{Explore the range $x \in [t\!-\!5\,\ldots\,t\!+\!5]$.}
    \IF{all $H_T(x) \le R\,{\cdot}\,(\mbox{number of pixels})$}
        \STATE{\textbf{break}}
    \ELSE
        \STATE{$t \gets t + 1$}
    \ENDIF
\ENDFOR
\STATE{\textbf{return} $t$.}
\end{algorithmic}
\end{algorithm}

Finally, note that assuming the cardinality of the finite discrete range for the threshold values is a constant, the histogram of thresholds could be built in linear time, that is, the computational complexity is linear in the number of central points. In computer-vision applications, residuals often take integral values between $0$ and $255$ inclusive, and the same is true for the thresholds, so the assumption is easily satisfied.

%although many authors look beyond that as well, e.g. \cite{Rosin98,Bradley07,Tao08}. Here we attempted to utilize certain spatial information in point's neighborhood keeping the computational cost at a low level. In fact, our approach could be seen as indirect estimation of optimal value at risk by mean of histogram of thresholds.

\subsection{Extensions}
\label{sec:ext}

Intuitively, it is clear that we are interested in the study of the tail of the random variable, whose histogram of thresholds we have constructed. 
That is: The region around the mode of the histogram (approximately $50\%$ of its area) mostly contains noise. We could search for the optimum threshold from the right-hand side of the region around the mode of the histogram of thresholds $H_T(t)$, until the rate of noisy patterns in Figure~\ref{fig:configs} falls below certain value.

One could also require the threshold to be above $2/3$ of residual values in the image. This requirement is typically well-matched with expected amount of moving points in surveillance video. 

One can consider other notions of isolation, which may have further benefits. In particular, we have considered $5 \times 5$ vicinity of each points and account only for those configurations where internal $3 \times 3$ pattern from Figure~\ref{fig:configs} is completely isolated from the pixels on outer border of $5 \times 5$ neighbourhood with values exceeding a threshold. In other words, one-pixel thin lines should not influence threshold estimation. However, the implementation would be more complicated in this case. We hence consider only the $3 \times 3$ patches.

%In our case, we compute the conditional value at risk of the probability mass function related over the thresholds, which is illustrated in Figure~\ref{fig:histthr}.

%Note that assuming the cardinality of the finite discrete range for the threshold values is a constant, the histogram of thresholds could be built in linear time, that is, the computational complexity is linear in the number of central points.
%(\textbf{TODO: do you want the word ``central''}). Yes.
%See Algorithm \ref{alg:thresholdingCVaR} for the pseudo-code.
%In computer-vision applications, residuals often take integral values between $0$ and $255$ inclusive, and the same is true for the thresholds, so the assumption is easily satisfied. 

%\begin{algorithm}[bht]
%\caption{{\tt threshold}}
%\label{alg:thresholdingCVaR}
%\algsetup{indent=2em}
%\begin{algorithmic}[1]
%\STATE{\textbf{Input}: integral residuals $r_i \in [0, 255]$, $\alpha$ such as 0.0025.}
%\STATE{\textbf{Output}:  threshold $t$.}
%\STATE{Initialise threshold $t$ to 255 and probability $p$ to 0.}
%\STATE{Build the histogram  $H_T(t)$ of thresholds from $r_i$.}
%\WHILE{$t > 1$}
%    \STATE{$p \leftarrow p + H_T(t)$ }
%    \IF{$p \ge \alpha$}
%        \STATE{\textbf{break}}
%    \ELSE
%        \STATE{$t \gets t + 1$}
%    \ENDIF
%\ENDWHILE
%\STATE{\textbf{return} $t$.}
%\end{algorithmic}
%\end{algorithm}

%-------------------------------------------------------------------------
\section{Experimental Evaluation}
\label{sec:exp}
%-------------------------------------------------------------------------
% (\url{http://changedetection.net})

To demonstrate our approach, we have implemented the autoencoder in Python~3 using \texttt{PyTorch}, the deep-learning engine. 
For evaluation and comparison against other methods, we present results on changedetection.net (cdnet~2014), a  well-established benchmark of \cite{Goyette12}. Although cdnet-2014 is not the benchmark where the  convolutional autoencoder would work best, considering  that the  video sequences in cdnet~2014 are quite short ($1,200$ to $9,000$ frames), we perform three passes over each sequence, in order to circumvent the shortage of data. Evaluation and computation of the foreground mask is performed only during the final pass. Note that we do not use any labelling information in the training at all, in sharp contrast to some other authors testing deep-learning approaches on cdnet-2014. For scoring, we utilize  cdnet~2014 evaluation software on categories: ``badWeather'', ``dynamicBackground'', ``cameraJitter'', ``baseline'', ``nightVideos'', ``shadow''. Each category contains several videos.
Since our autoencoder expects a standard image size of $576{\times}704$, we resize input image, if necessary, and the computed foreground mask is resized back to the original size. This procedure affects both quality and performance, but seems unavoidable because network architecture is determined by the image size.

For comparison, we have chosen several methods from \texttt{LRSLibrary}, an excellent toolbox developed by \mbox{A.~Sobral} and co-authors \cite{lrslibrary2015, bouwmans2015}.
Considering that our approach can be seen as a low-rank approximation method, we have focussed on five well-performing  methods considering the ``low-rank and sparse'' model for background modeling and subtraction in videos. The five methods were: \texttt{LRR\_FastLADMAP} \cite{Lin11}, \texttt{MC\_GROUSE} \cite{Balzano13}, \texttt{RPCA\_FPCP} \cite{Rodriguez13}, \texttt{ST\_GRASTA} \linebreak \cite{He11}, \texttt{TTD\_3WD} \cite{Oreifej13}, when considering both statistical performance and the run-time per frame. 
%based on the number of citations, year of publication, and run-time performance. The latter is important because many advanced algorithms are painfully slow and as such impractical.}{based primarily on run-time performance because many advanced algorithms are painfully slow.}
We also used the recent and state-of-the-art algorithm \texttt{OMoGMF}, proposed in \cite{Meng13,Yong18}, as implemented in Matlab by the authors. \texttt{OMoGMF} was identified as a top performer in our experiments. The only caveat to keep in mind is that \texttt{OMoGMF} algorithm outputs black foreground mask after about $2,500$ frames. As a workaround, we discard the foreground masks with all black pixels from the scoring process.
Considering neither of the methods used for comparison requires a fixed frame size, we do not rescale the frames.

%Experiments revealed that many algorithms experience difficulties with challenging videos in cdnet~2014. In particular, computation of foreground mask is a weak point in many implementations at least in the standard setup provided by \cite{lrslibrary2015}.

\begin{table*}[th]
\begin{center}
{\footnotesize
\begin{tabular}{|l|c|c|c|c|c|c|S|}
\hline
\textit{Method} & \textit{Recall} & \textit{Specificity} & \textit{FPR}
& \textit{FNR} & \textit{Precision} & \textit{F1} & {Run-time \;[s/frame]} \\
\hline
LRR\_FastLADMAP \cite{Lin11}        & 0.74694 & 0.93980 & 0.06020 & 0.25306 & 0.28039 & 0.36194 & 4.611  \\
MC\_GROUSE \cite{Balzano13}         & 0.65640 & 0.89692 & 0.10308 & 0.34360 & 0.25425 & 0.31495 & 10.621 \\
OMoGMF \cite{Meng13,Yong18}         & 0.89943 & 0.98289 & 0.01711 & 0.10057 & 0.62033 & 0.72611 & 0.123  \\
RPCA\_FPCP \cite{Rodriguez13}       & 0.73848 & 0.94733 & 0.05267 & 0.26152 & 0.29994 & 0.37900 & 0.504  \\
ST\_GRASTA \cite{He11}              & 0.45340 & 0.98205 & 0.01795 & 0.54660 & 0.44009 & 0.42367 & 3.266  \\
TTD\_3WD \cite{Oreifej13}           & 0.61103 & 0.97117 & 0.02883 & 0.38897 & 0.35557 & 0.40297 & 10.343 \\
Autoencoder, $5{\times}5$ min. thr. & 0.65676 & 0.99360 & 0.00640 & 0.34324 & 0.77756 & 0.70593 & 0.57   \\
\hline
\end{tabular}}
\end{center}
\caption{Performance results on ``baseline'' video-category from \url{http://changedetection.net}. The execution time in seconds per frame is given for ``highway'' video-sequence with $240{\times}320$ images. The last line reflects only forward/backward single iteration time.}.
\label{tab:all-baseline}
%\vspace{-20pt}
\end{table*}

% Long and short paper versions are different here.

The data matrix has been built from $50$ most recent frames, which is the default setting used across both  \texttt{LRSLibrary} and \texttt{OMoGMF}.
Upon arrival of a new image, it was inserted at the end of the queue keeping the time-ordering, which some algorithms might be sensitive to. 

\subsection{Statistical Performance}

Results for the ``baseline'' category, along with timing information in the last column for ``Highway'' video, are presented in Table~\ref{tab:all-baseline}.

%\subsection{Our Implementation}

%The most heavy part of computation had been done on GPU side.  

Table~\ref{tab:all-six-cat} summarizes scoring results obtained on $6$ video-categories for all the methods including \texttt{OMoGMF}, which was identified as the best subspace method in this study, and our convolutional autoencoder. Note, there are two sub-tables for autoencoder results. In the first case (``min. threshold'') we derive the optimum threshold from a distribution of minimum residuals across a batch of reconstructed backgrounds, but \textit{without} looking at points' neighbourhoods. In the second case (``$5{\times}5$ min. threshold''), threshold was computed with the extensions described in the previous section, formula (\ref{eq:residvalue}), and that leads to an improvement in the overall $F$-measure, especially for dynamic-background category.

\begin{figure*}[th]
\begin{center}
\includegraphics[width=\textwidth,height=0.11\textheight]{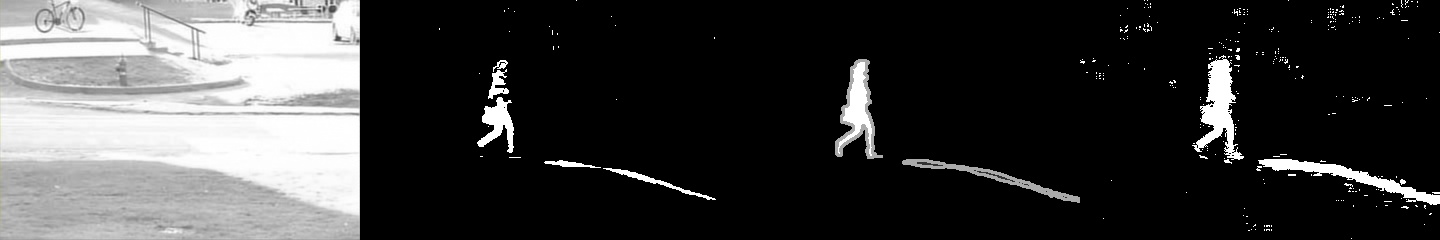}
\vspace{1pt}
\includegraphics[width=\textwidth,height=0.11\textheight]{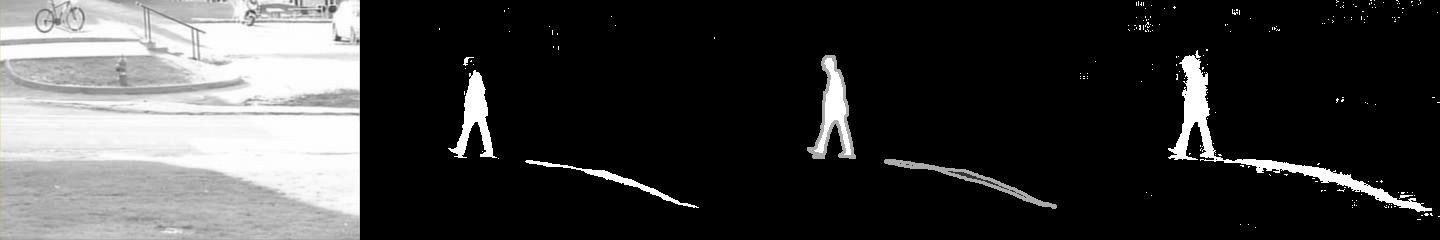}
\vspace{1pt}
\includegraphics[width=\textwidth,height=0.11\textheight]{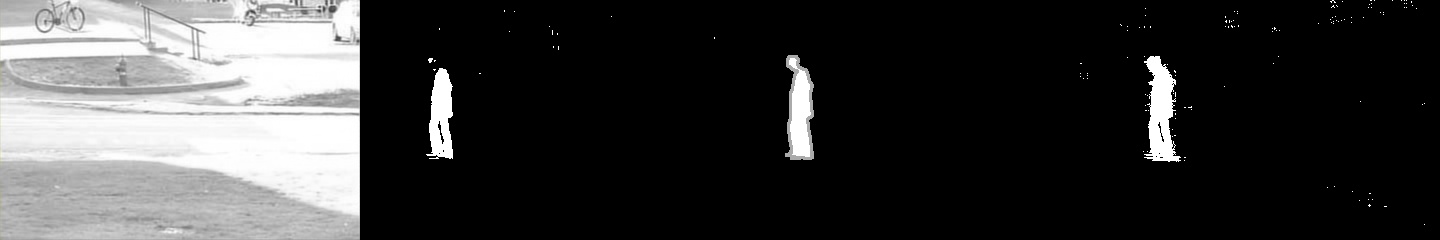}
\vspace{1pt}
\includegraphics[width=\textwidth,height=0.11\textheight]{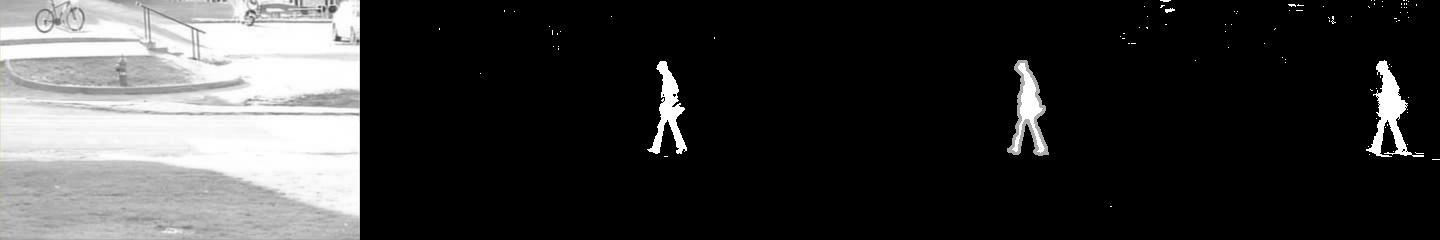}
\vspace{1pt}
\includegraphics[width=\textwidth,height=0.11\textheight]{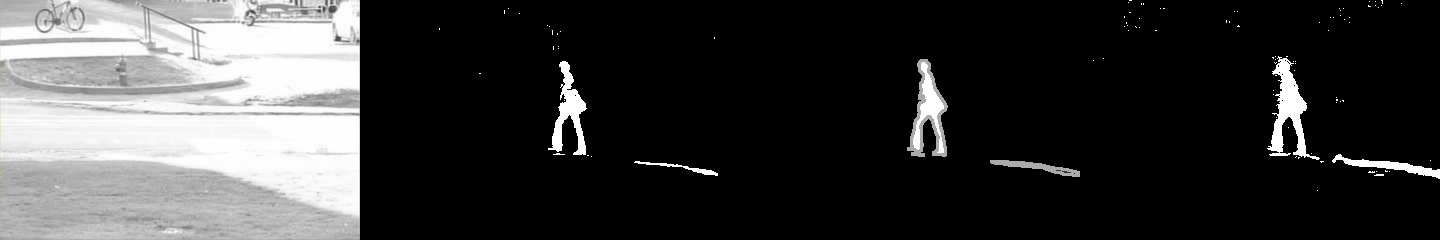}
\vspace{1pt}
\includegraphics[width=\textwidth,height=0.11\textheight]{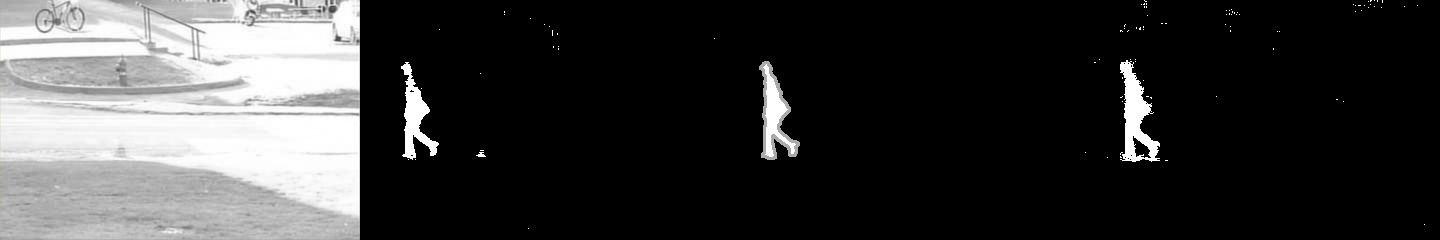}
\end{center}
\vspace{-5pt}
\caption{``Pedestrians'' video-sequence, cdnet~2014. \textit{Left to right}: reconstructed background outputted by the \texttt{Autoencoder}, foreground mask by the \texttt{Autoencoder}, the ground-truth, and foreground mask obtained by \texttt{OMoGMF}. Foreground mask obtained from autoencoder output does not undergo any post-processing.}
\vspace{-10pt}
\label{fig:pedestrians}
\end{figure*}

\begin{figure*}[th]
\begin{center}
\includegraphics[width=\textwidth,height=0.125\textheight]{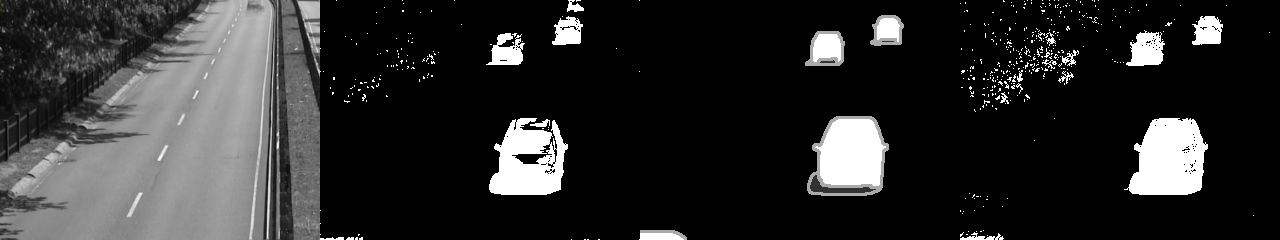}
\vspace{1pt}
\includegraphics[width=\textwidth,height=0.125\textheight]{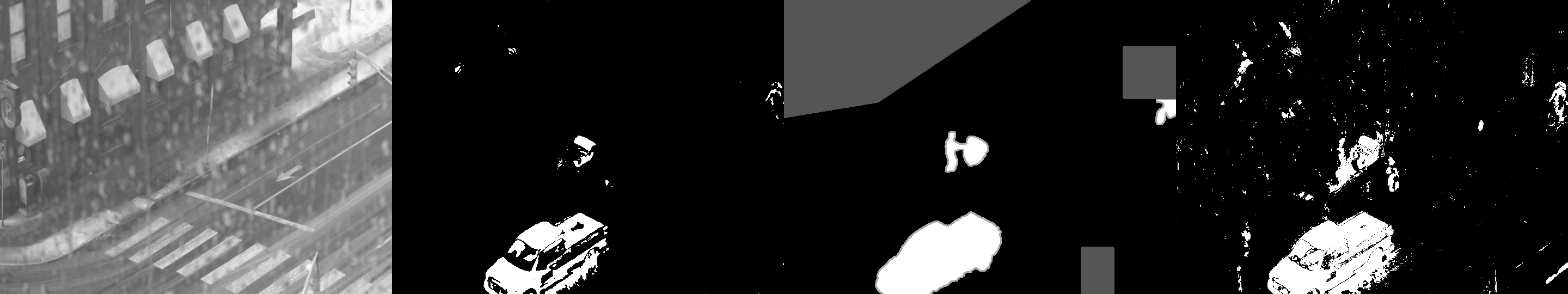}
\vspace{1pt}
\includegraphics[width=\textwidth,height=0.125\textheight]{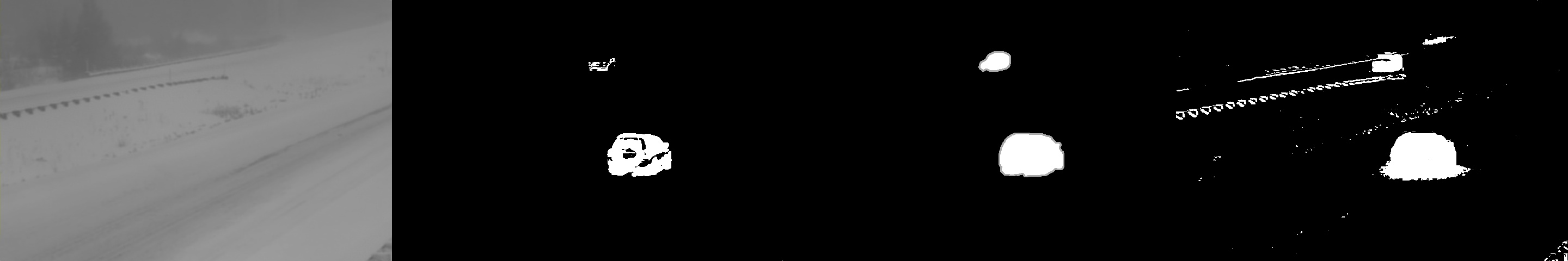}
\vspace{1pt}
\includegraphics[width=\textwidth,height=0.125\textheight]{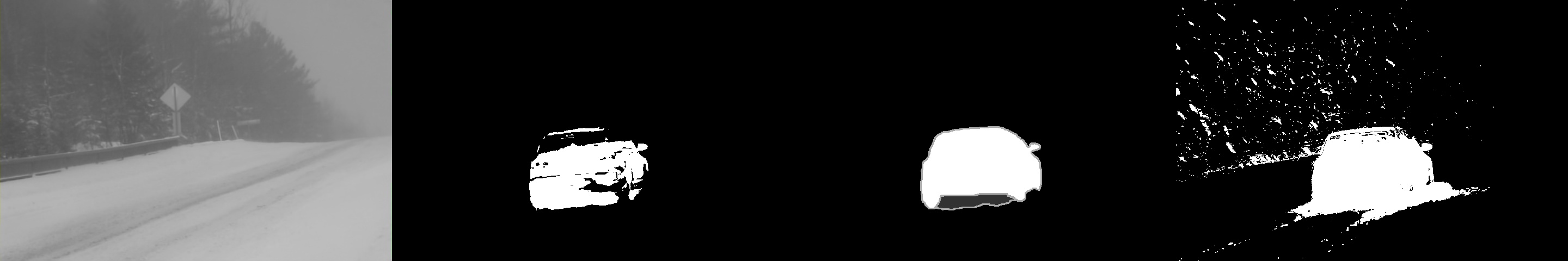}
\vspace{1pt}
\includegraphics[width=\textwidth,height=0.125\textheight]{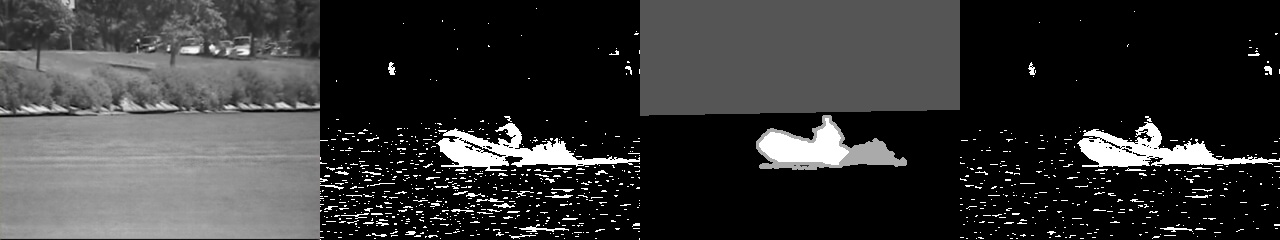}
\vspace{1pt}
\includegraphics[width=\textwidth,height=0.125\textheight]{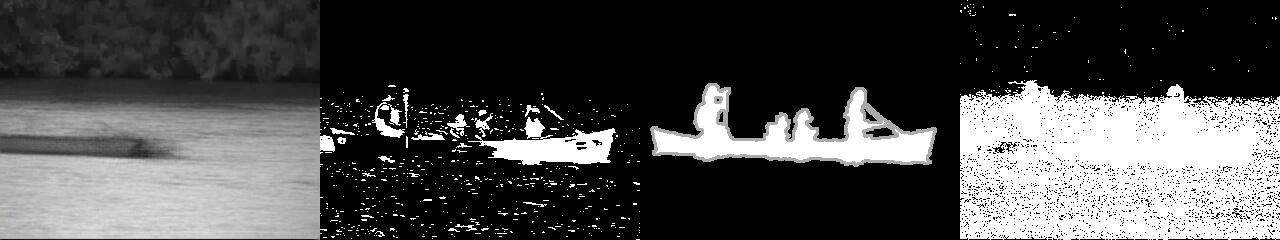}
\vspace{1pt}
\includegraphics[width=\textwidth,height=0.125\textheight]{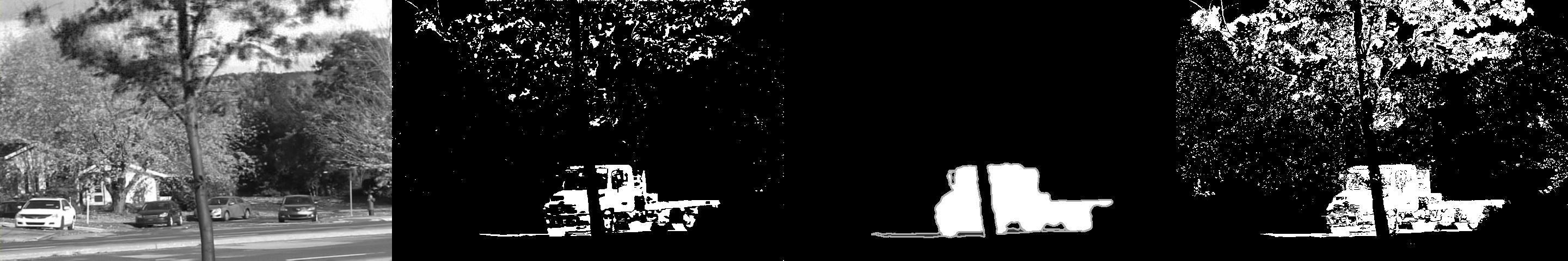}
\end{center}
\vspace{-5pt}
\caption{Selected background subtraction results. \textit{Left to right}: reconstructed background outputted by the \texttt{Autoencoder}, foreground mask by the \texttt{Autoencoder}, the ground-truth, and foreground mask obtained by \texttt{OMoGMF}. Video-sequences, cdnet~2014, \textit{top to bottom}: ``highway'', ``wetsnow'', ``blizzard'', ``snowfall'', ``boats'', ``canoe'', ``fall''. Foreground mask obtained from autoencoder output does not undergo any post-processing.}
\label{fig:results}
\end{figure*}

Figures~\ref{fig:pedestrians} and ~\ref{fig:results} present selected results obtained by \texttt{Autoencoder} and \texttt{OMoGMF} methods. Here we provide some typical cases, where advantages and disadvantages of both approaches can be clearly seen. A few observations deserve attention. First, \texttt{OMoGMF} performs really well in general. It integrates a flexible Gaussian Mixture Model (GMM) and produces a solid motion mask, but it is less resistant to noise and non-stationary background than our  autoencoder. Second, our autoencoder yields a good background estimation and copes better with dynamic background than many other methods. While a single choice of a threshold is not flexible enough, as it produces  ``holes'' in the motion mask and worse scores, our VaR-based method seems more robust than the alternatives. %We are currently investigating several strategies for optimal thresholding. 
Third, foreground sometimes ``leaks'' into background, as for example in the  ``canoe'' result in Figure~\ref{fig:results}. Partially, this can be explained by videos being too short for proper training of autoencoder, which tends to memorize images --- the known problem. Also note that autoencoder operates on grayscale images, unlike other methods. By taking advantage of colour images the detectability of moving objects can be improved, whereas realistic scenarios include night-time videos, where colour information is not available anyway.

\subsection{Runtime}

All the methods used for comparison, except our  autoencoder, are implemented in Matlab. It took several weeks to process selected videos on Intel Core i7-4800MQ, 4-core CPU, 16~Gb, 2.70~GHz workstation powered by RedHat~7.6/64 Linux and Matlab~2018a.
To make use of the general-purpose graphics processing unit (GPGPU), we have utilised a different machine  to run the autoencoder. This machine was equipped with Intel Xeon E5-2699 CPU at 2.20~GHz and Tesla K40c GPGPU with 12 GB of on-board memory, and ran by RedHat Linux 7.5. 
With the GPGPU, it took about $2$ days to run on the benchmark.

Since we are using different hardware for our autoencoder and \texttt{OMoGMF}, it is not straightforward to compare their computational speed.
In general, our autoencoder utilizes a single CPU core and the GPGPU specified above. \texttt{OMoGMF} utilizes $4$ CPU cores and takes about $0.123$ seconds per a $240{\times}320$ frame (video-sequence ``highway''). 
In the case of our autoencoder, the DIS algorithm followed by forward/backward steps take\Highlit{s} about $0.57$ sec. per a $576{\times}704$ frame. Thresholding and storing on hard-disk (for subsequent scoring) take about $1$ second. Our thresholding procedure has very simple code and can be easily improved to real-time performance. The actual bottleneck resides in neural network implementation. Considering $576{\times}704$ images and at least linear dependency of processing time on the number of pixels, the processing times of \texttt{OMoGMF} and the main part of  our autoencoder should be comparable.

%-------------------------------------------------------------------------
\section{A Summary and Discussion}
\label{sec:conclude}
%-------------------------------------------------------------------------

We have presented an algorithm for tracking of time-varying low-rank background models of time-varying matrices, using a continuously trained and applied convolutional autoencoder. Our approach displays solid performance overall, and seems comparable to the best subspace methods.  Three features may be worth highlighting. First, incremental training  makes the network adaptive to gradual scene changes,  which always happen in reality. Second, no labelled data is needed, in contrast to typical deep-learning approaches. Instead, we down-weighing the moving points using a rough estimation of the foreground mask. The training is primarily driven by background points and robust to outliers. Third, we compute foreground mask of moving objects by considering the spatial neighbourhood of each pixel and VaR-based thresholding.

% Long and short paper versions are different at the end of this paragraph.
Low-rank methods have made remarkable progress in recent years, but still  demonstrate certain limitations. As it turns out, all the methods considered in this study have difficulties in producing a convincing foreground motion mask in the case of fast-varying background. Clearly, there are good statistical and complexity-theoretic reasons \cite{whittle1988restless,gittins1989multi} for any method to have difficulties in fast-changing environments, but there may be a scope for improvement.
Improving upon thresholding technique can alleviate the problem of a poor foreground mask, to some extent. 
%The latter is often very slow and does not conform real-time requirement. The right step was taken in \cite{Yong18} where thresholding is integrated with background estimation.

Originally, we designed the algorithm to handle video streams, collected from a network of CCTV cameras in Dublin, Ireland, which posed a serious challenge to any algorithm we tested: camera jitter, compression artefacts, poor image quality, adverse weather condition, night videos and so on. Usually, it takes approximately $20$ minutes ($30,000$ video-frames) for autoencoder to build up a good background model. However, the standard video sequences, adopted by the community for benchmarking, are typically very short -- few thousands frames. This rises a question about the scoring process, particularly when algorithms based on deep learning architecture are involved. 
%While some algorithms demonstrate outstanding results with $F$-measure close to 1, it is unclear how they will they perform on a video-stream with million images (just one day of streaming data)? 
%In realistic scenario, when thousands of cameras to be deployed and supported, any training process would make a lot of pain. As such, explicit training phase should be completely avoided, if we are talking about practical applications. On the other hand, the baseline autoencoder, despite good results comparing to existing approaches, yields quite low $F$-measure overall. Using pre-trained network as in \cite{Lim18}, but without explicit training stage upon deployment, in our opinion, is a promising research direction. 

It seems that the next generation of benchmarks should be designed. This can be, for example, a collection of compressed, one-day long videos with few thousand check-point images evenly scattered across the sequence and manually labelled. When video-processing reaches the next check-point frame, the scoring procedure is applied to collect and update a performance statistics. That would offer a more realistic comparison protocol.

%\Highlit{We could say about VAE and GAN, but again, those NN is difficult to train and much more data is needed, hence better benchmarks.}

An important avenue for further efforts, as can be seen in the last column of Table~\ref{tab:all-baseline}, is the speed-up of the autoencoder. This is a principal limitation of any deep-learning  architecture, overall. Although hardware advances lessen the cost and increases the performance of GPGPUs at remarkable pace,  having a GPGPU per camera may still be too expensive for many applications. One possible solution is to run forward steps often and backward (training) ones rarely. This will reduce the rate of adaptation, but also decrease the run-time.

\bibliographystyle{ieeetr}
\bibliography{egbib}

\FloatBarrier

\begin{table*}[th]
\begin{center}
{\small
\begin{tabular}{|l|c|c|c|c|c|c|}
\hline
\textit{Video} & \textit{Recall} & \textit{Specificity} & \textit{FPR}
& \textit{FNR} & \textit{Precision} & \textit{F1}     \\
\hline
\textbf{LRR\_FastLADMAP \cite{Lin11}}: & {} & {} & {} & {} & {} & {} \\
badWeather        & 0.82941 & 0.82644 & 0.17356 & 0.17059 & 0.09239 & 0.15580 \\
baseline          & 0.74694 & 0.93980 & 0.06020 & 0.25306 & 0.28039 & 0.36194 \\
cameraJitter      & 0.75423 & 0.83766 & 0.16234 & 0.24577 & 0.18119 & 0.28715 \\
dynamicBackground & 0.69953 & 0.79853 & 0.20147 & 0.30047 & 0.03828 & 0.06968 \\
nightVideos       & 0.80056 & 0.84435 & 0.15565 & 0.19944 & 0.11062 & 0.18503 \\
shadow            & 0.72950 & 0.88521 & 0.11479 & 0.27050 & 0.23030 & 0.32793 \\
\hline
Overall           & 0.76003 & 0.85533 & 0.14467 & 0.23997 & 0.15553 & \textbf{0.23125} \\
\hline
\hline
\textbf{ST\_GRASTA \cite{He11}}: & {} & {} & {} & {} & {} & {} \\
badWeather        & 0.26555 & 0.98971 & 0.01029 & 0.73445 & 0.45526 & 0.30498 \\
baseline          & 0.45340 & 0.98205 & 0.01795 & 0.54660 & 0.44009 & 0.42367 \\
cameraJitter      & 0.51138 & 0.91313 & 0.08687 & 0.48862 & 0.23995 & 0.31572 \\
dynamicBackground & 0.41411 & 0.94755 & 0.05245 & 0.58589 & 0.08732 & 0.13736 \\
nightVideos       & 0.42488 & 0.97224 & 0.02776 & 0.57512 & 0.24957 & 0.28154 \\
shadow            & 0.44317 & 0.96681 & 0.03319 & 0.55683 & 0.42604 & 0.41515 \\
\hline
Overall           & 0.41875 & 0.96192 & 0.03808 & 0.58125 & 0.31637 & \textbf{0.31307} \\
\hline
\hline
\textbf{RPCA\_FPCP \cite{Rodriguez13}}: & {} & {} & {} & {} & {} & {} \\
badWeather        & 0.82546 & 0.84424 & 0.15576 & 0.17454 & 0.09950 & 0.16687 \\
baseline          & 0.73848 & 0.94733 & 0.05267 & 0.26152 & 0.29994 & 0.37900 \\
cameraJitter      & 0.74452 & 0.84143 & 0.15857 & 0.25548 & 0.18436 & 0.29024 \\
dynamicBackground & 0.69491 & 0.80688 & 0.19312 & 0.30509 & 0.03928 & 0.07134 \\
nightVideos       & 0.79284 & 0.85751 & 0.14249 & 0.20716 & 0.11797 & 0.19497 \\
shadow            & 0.72132 & 0.90454 & 0.09546 & 0.27868 & 0.26474 & 0.36814 \\
\hline
Overall           & 0.75292 & 0.86699 & 0.13301 & 0.24708 & 0.16763 & \textbf{0.24509} \\
\hline
\hline
\textbf{OMoGMF \cite{Meng13,Yong18}}: & {} & {} & {} & {} & {} & {} \\
badWeather        & 0.86871 & 0.98939 & 0.01061 & 0.13129 & 0.57917 & 0.67214 \\
baseline          & 0.89943 & 0.98289 & 0.01711 & 0.10057 & 0.62033 & 0.72611 \\
cameraJitter      & 0.85954 & 0.90739 & 0.09261 & 0.14046 & 0.30566 & 0.44235 \\
dynamicBackground & 0.87655 & 0.86383 & 0.13617 & 0.12345 & 0.08601 & 0.15012 \\
nightVideos       & 0.75607 & 0.92372 & 0.07628 & 0.24393 & 0.23252 & 0.31336 \\
shadow            & 0.55771 & 0.80276 & 0.03057 & 0.27562 & 0.40539 & 0.37449 \\
\hline
Overall           & 0.80300 & 0.91166 & 0.06056 & 0.16922 & 0.37151 & \textbf{0.44643} \\
\hline
\hline
\textbf{Autoencoder, min. threshold}: & {} & {} & {} & {} & {} & {} \\
badWeather        & 0.83978 & 0.97169 & 0.02831 & 0.16022 & 0.46847 & 0.57446 \\
baseline          & 0.72216 & 0.98873 & 0.01127 & 0.27784 & 0.62647 & 0.63878 \\
cameraJitter      & 0.74510 & 0.92411 & 0.07589 & 0.25490 & 0.40775 & 0.50148 \\
dynamicBackground & 0.81096 & 0.86072 & 0.13928 & 0.18904 & 0.08430 & 0.14604 \\
nightVideos       & 0.66925 & 0.95927 & 0.04073 & 0.33075 & 0.23837 & 0.34159 \\
shadow            & 0.71458 & 0.98464 & 0.01536 & 0.28542 & 0.69253 & 0.68589 \\
\hline
Overall           & 0.75031 & 0.94819 & 0.05181 & 0.24969 & 0.41965 & \textbf{0.48137} \\
\hline
\hline
\textbf{Autoencoder, $5{\times}5$ min. thr.}: & {} & {} & {} & {} & {} & {} \\
badWeather        & 0.64712 & 0.99898 & 0.00102 & 0.35288 & 0.92511 & 0.75683 \\
baseline          & 0.65676 & 0.99360 & 0.00640 & 0.34324 & 0.77756 & 0.70593 \\
cameraJitter      & 0.64940 & 0.95970 & 0.04030 & 0.35060 & 0.43331 & 0.51465 \\
dynamicBackground & 0.55339 & 0.97154 & 0.02846 & 0.44661 & 0.24313 & 0.30643 \\
nightVideos       & 0.53979 & 0.97544 & 0.02456 & 0.46021 & 0.33687 & 0.39017 \\
shadow            & 0.62316 & 0.98857 & 0.01143 & 0.37684 & 0.76103 & 0.67776 \\
\hline
Overall           & 0.61160 & 0.98130 & 0.01870 & 0.38840 & 0.57950 & \textbf{0.55863} \\
\hline
\end{tabular}}
\end{center}
\caption{Performance results on 6 video-categories from \url{http://changedetection.net}. For autoencoder we present results for two thresholding strategies in the last two sub-tables as detailed in Section~\ref{sec:algo}.}
\label{tab:all-six-cat}
\end{table*}

\begin{figure*}[th]
\begin{center}
\includegraphics[width=\textwidth,height=0.09\textheight]{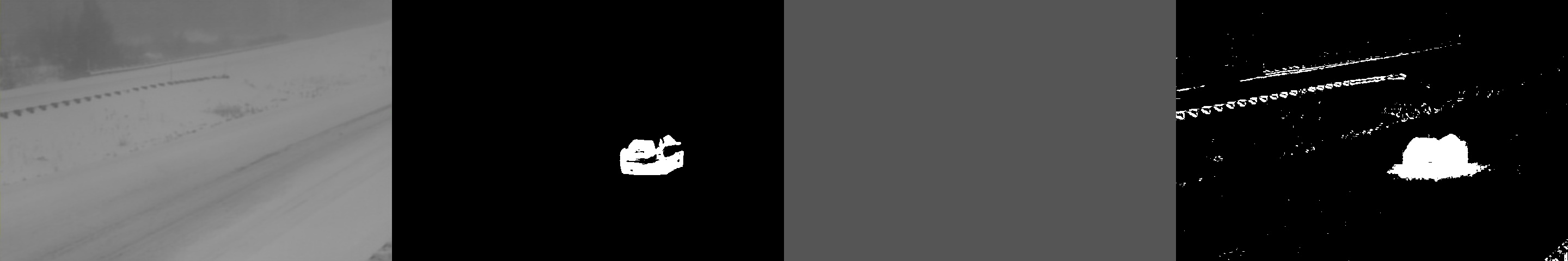}
\vspace{1pt}
\includegraphics[width=\textwidth,height=0.09\textheight]{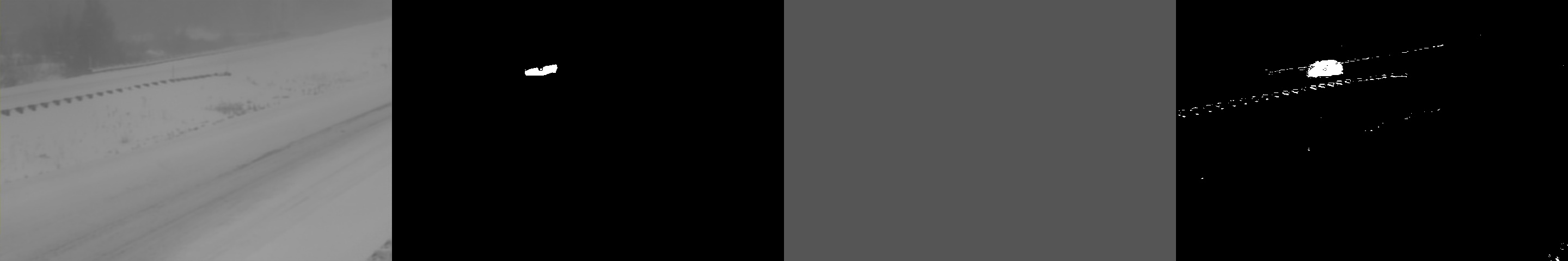}
\vspace{1pt}
\includegraphics[width=\textwidth,height=0.09\textheight]{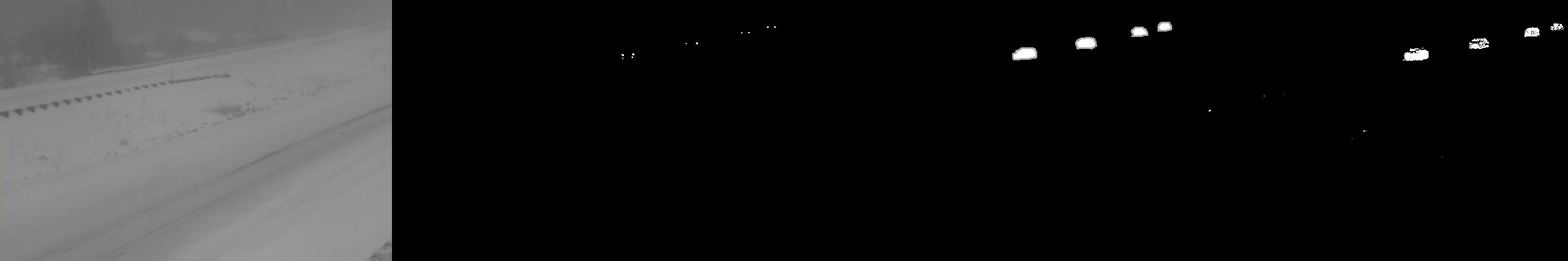}
\vspace{1pt}
\includegraphics[width=\textwidth,height=0.09\textheight]{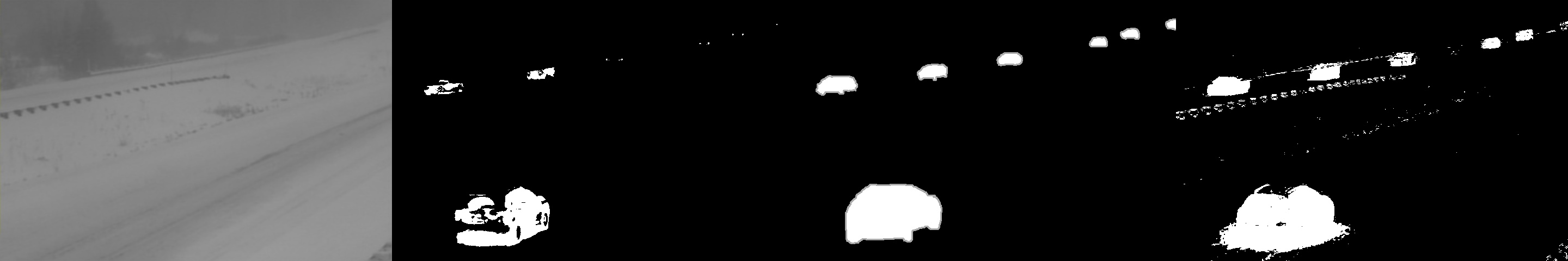}
\vspace{1pt}
\includegraphics[width=\textwidth,height=0.09\textheight]{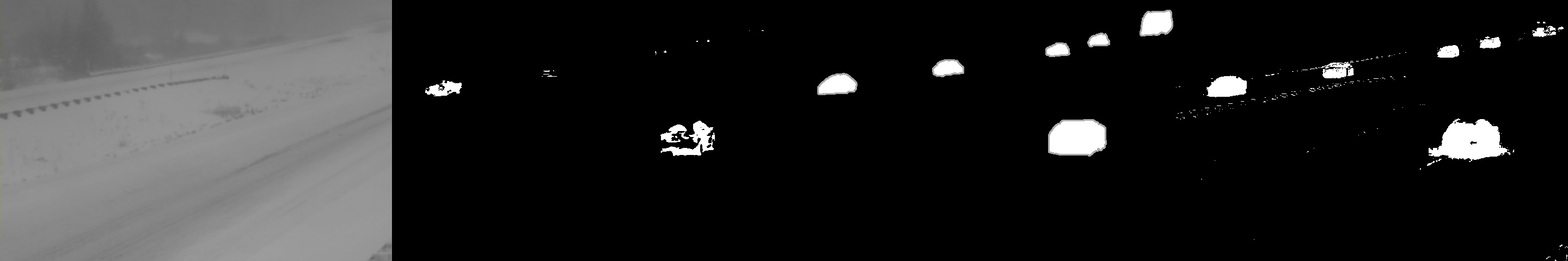}
\vspace{1pt}
\includegraphics[width=\textwidth,height=0.09\textheight]{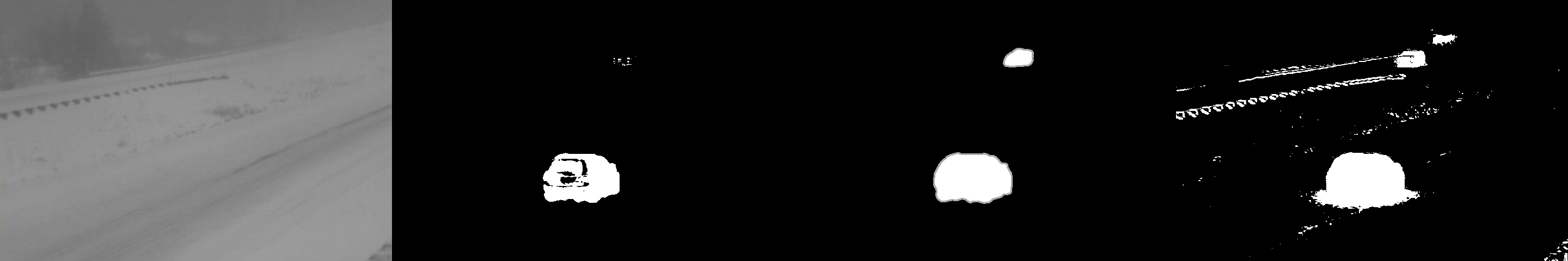}
\vspace{1pt}
\includegraphics[width=\textwidth,height=0.09\textheight]{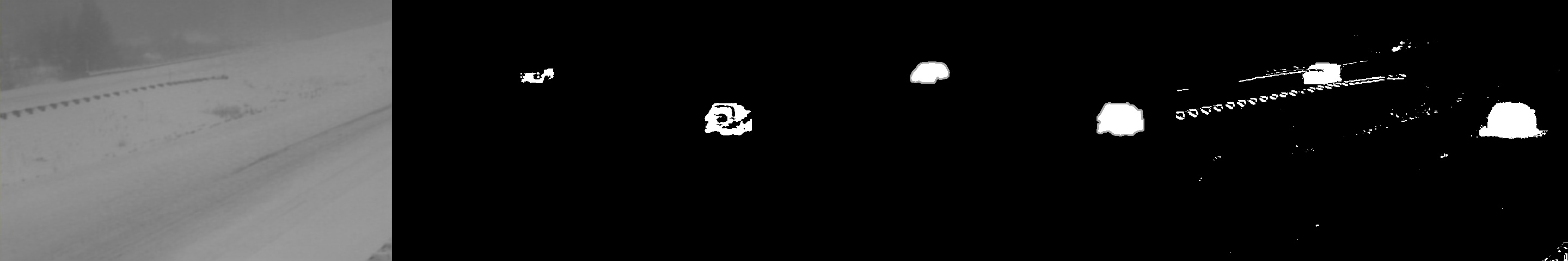}
\vspace{1pt}
\includegraphics[width=\textwidth,height=0.09\textheight]{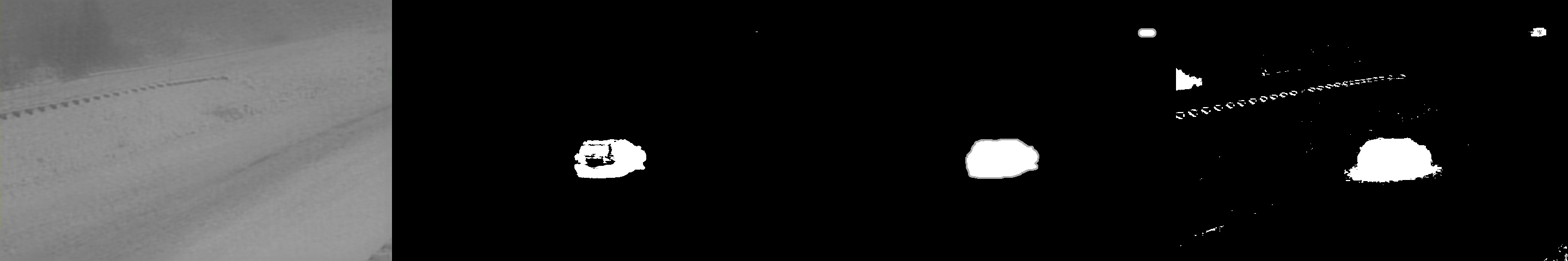}
\vspace{1pt}
\includegraphics[width=\textwidth,height=0.09\textheight]{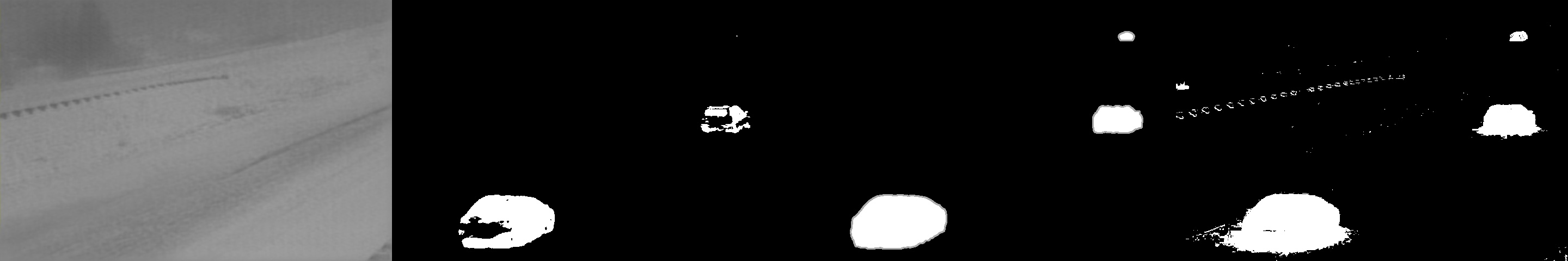}
\vspace{1pt}
\includegraphics[width=\textwidth,height=0.09\textheight]{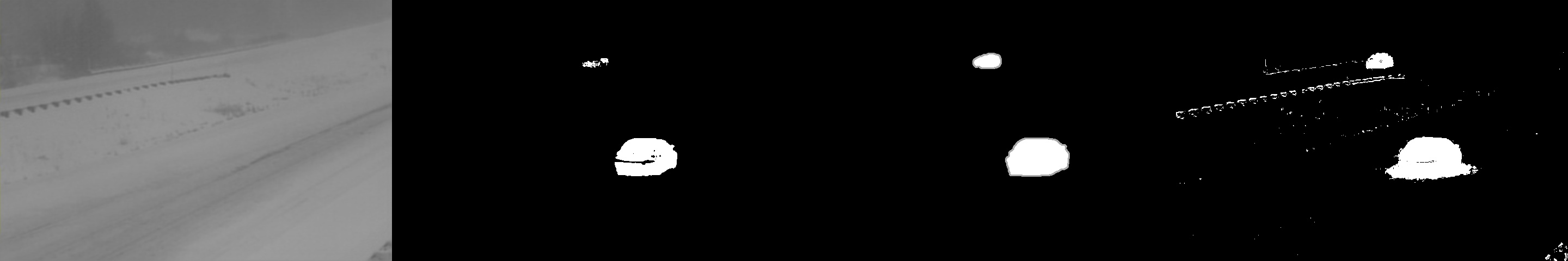}
\end{center}
\vspace{-5pt}
\caption{``Blizzard'' video-sequence, cdnet~2014. \textit{Left to right}: reconstructed background outputted by the \texttt{Autoencoder}, foreground mask by the \texttt{Autoencoder}, the ground-truth, and foreground mask obtained by \texttt{OMoGMF}. \texttt{OMoGMF} is good in detecting small objects,  but noise is also often acquired. Foreground mask obtained from autoencoder output does not undergo any post-processing.}
\label{fig:blizzard}
\end{figure*}
\begin{figure*}[th]
\begin{center}
\includegraphics[width=\textwidth,height=0.11\textheight]{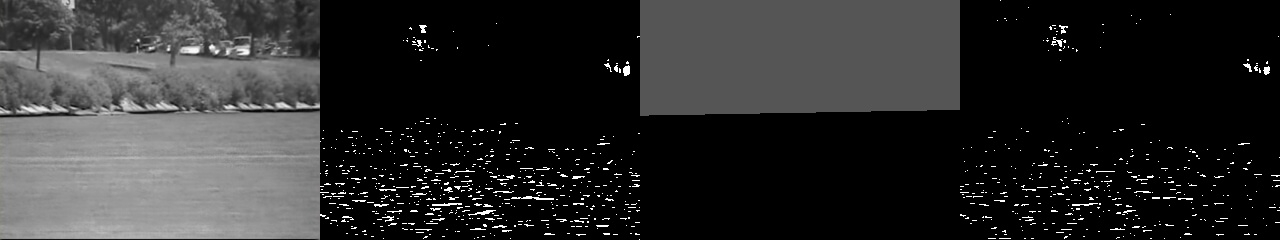}
\vspace{1pt}
\includegraphics[width=\textwidth,height=0.11\textheight]{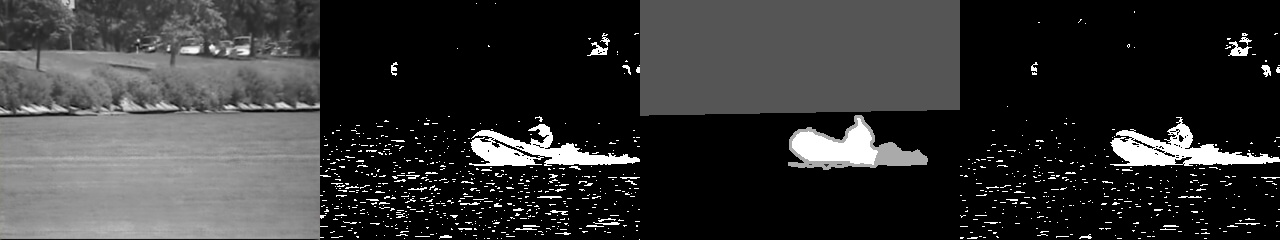}
\vspace{1pt}
\includegraphics[width=\textwidth,height=0.11\textheight]{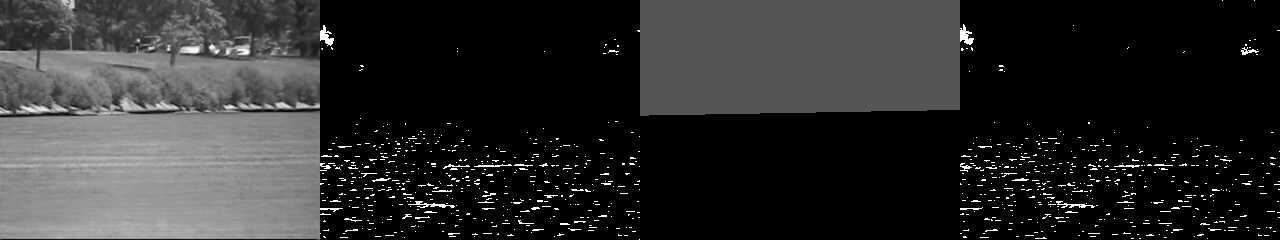}
\vspace{1pt}
\includegraphics[width=\textwidth,height=0.11\textheight]{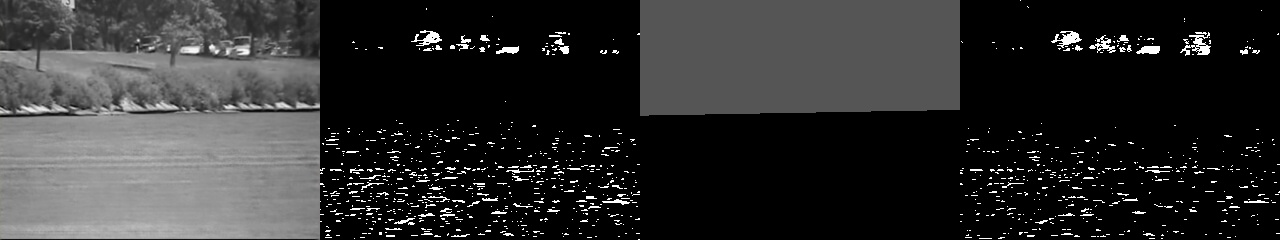}
\vspace{1pt}
\includegraphics[width=\textwidth,height=0.11\textheight]{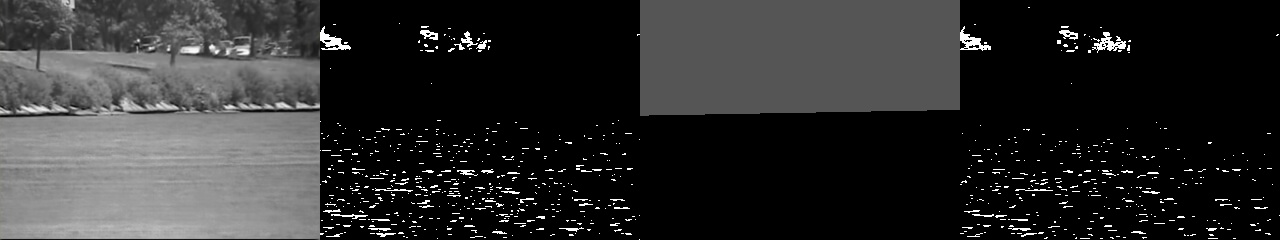}
\vspace{1pt}
\includegraphics[width=\textwidth,height=0.11\textheight]{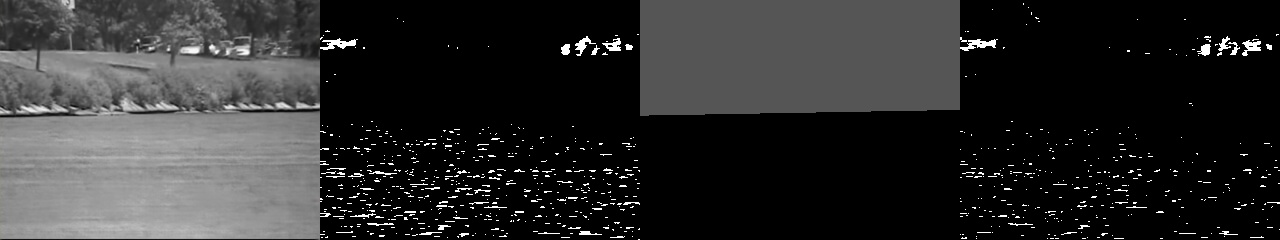}
\vspace{1pt}
\includegraphics[width=\textwidth,height=0.11\textheight]{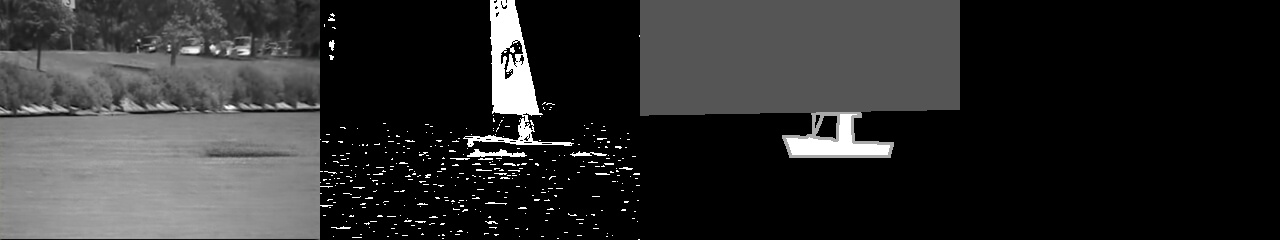}
\vspace{1pt}
\includegraphics[width=\textwidth,height=0.11\textheight]{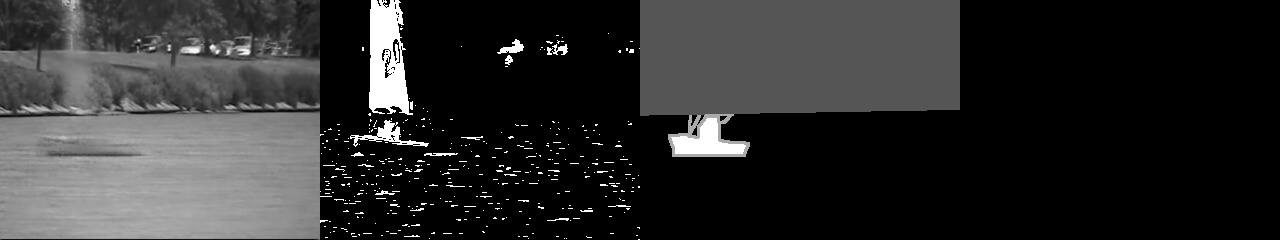}
\end{center}
\vspace{-5pt}
\caption{``Boats'' video-sequence, cdnet~2014. \textit{Left to right}: reconstructed background outputted by the \texttt{Autoencoder}, foreground mask by the \texttt{Autoencoder}, the ground-truth, and foreground mask obtained by \texttt{OMoGMF}. Non-stationary background is a particularly difficult case for modelling by any method mentioned in this study. Note that there is an issue in \texttt{OMoGMF} code that results in black foreground mask after approximately $2,500$ frames. Foreground mask obtained from autoencoder output does not undergo any post-processing. Some ``leakage'' of slowly moving objects into reconstructed background can be observed in the last two rows.}
\label{fig:boats}
\end{figure*}
\begin{figure*}[th]
\begin{center}
\includegraphics[width=\textwidth,height=0.11\textheight]{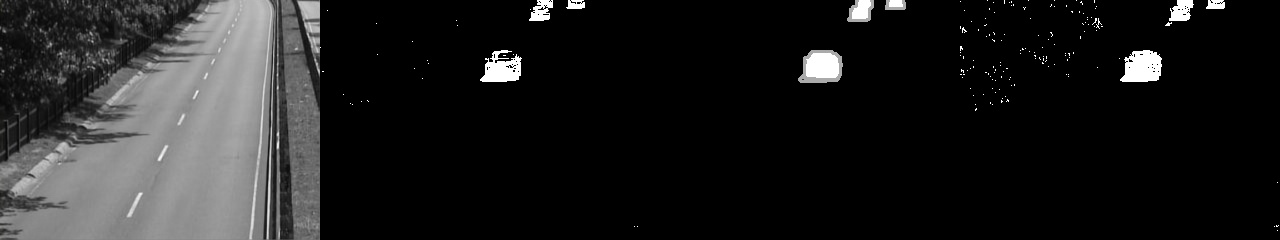}
\vspace{1pt}
\includegraphics[width=\textwidth,height=0.11\textheight]{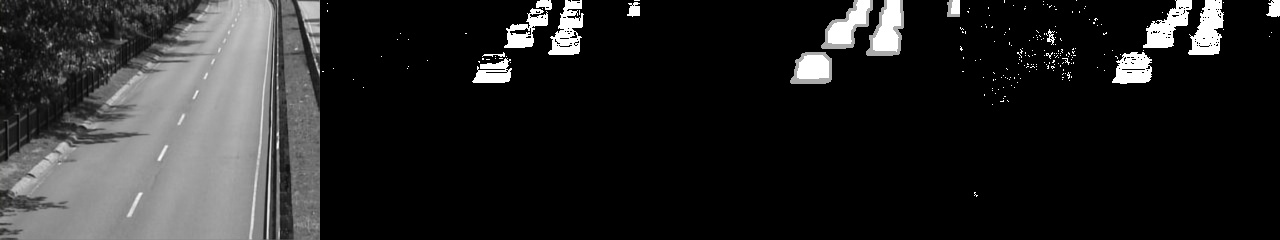}
\vspace{1pt}
\includegraphics[width=\textwidth,height=0.11\textheight]{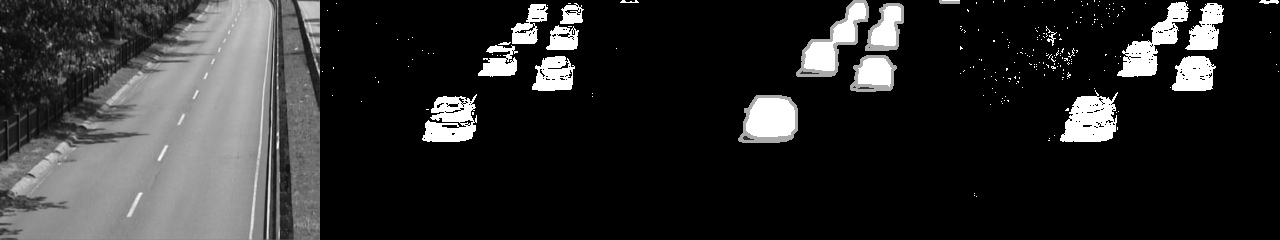}
\vspace{1pt}
\includegraphics[width=\textwidth,height=0.11\textheight]{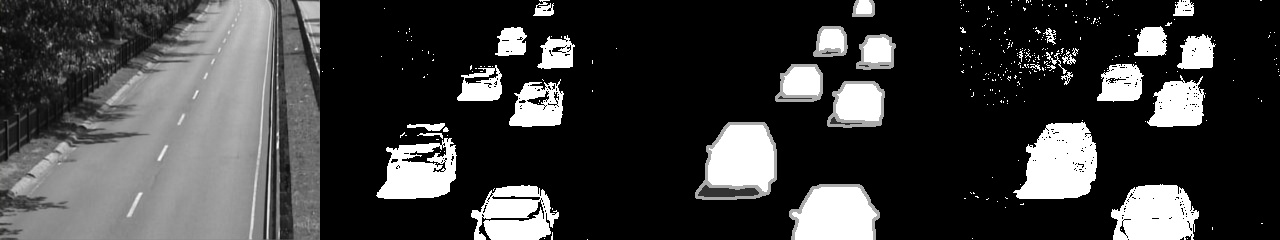}
\vspace{1pt}
\includegraphics[width=\textwidth,height=0.11\textheight]{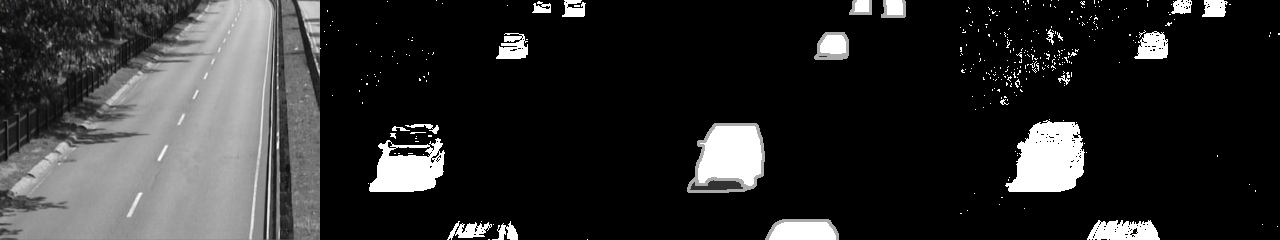}
\vspace{1pt}
\includegraphics[width=\textwidth,height=0.11\textheight]{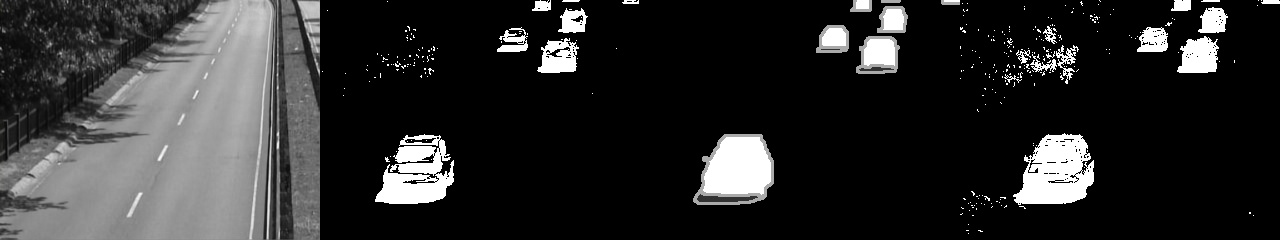}
\vspace{1pt}
\includegraphics[width=\textwidth,height=0.11\textheight]{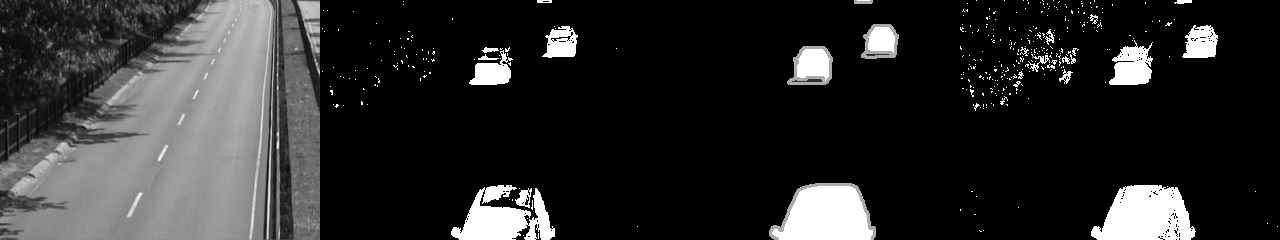}
\vspace{1pt}
\includegraphics[width=\textwidth,height=0.11\textheight]{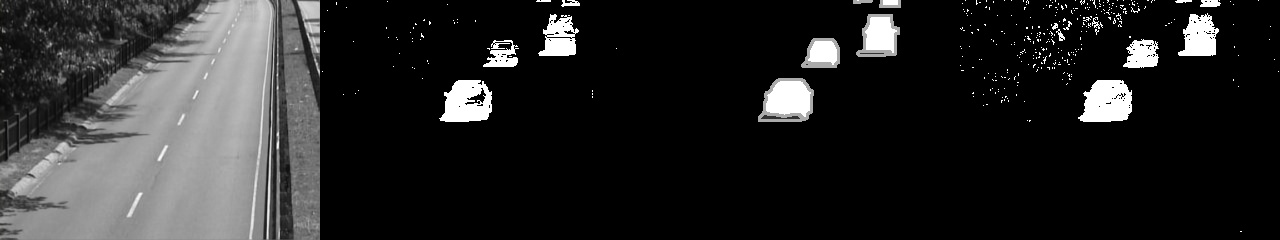}
\end{center}
\vspace{-5pt}
\caption{``Highway'' video-sequence, cdnet~2014. \textit{Left to right}: reconstructed background outputted by the \texttt{Autoencoder}, foreground mask by the \texttt{Autoencoder}, the ground-truth, and foreground mask obtained by \texttt{OMoGMF}. Notice that the autoencoder is more resistant against dynamic background, at the expense of less solid motion mask. Foreground mask obtained from autoencoder output does not undergo any post-processing.}
\label{fig:highway}
\end{figure*}
\begin{figure*}[th]
\begin{center}
\includegraphics[width=\textwidth,height=0.11\textheight]{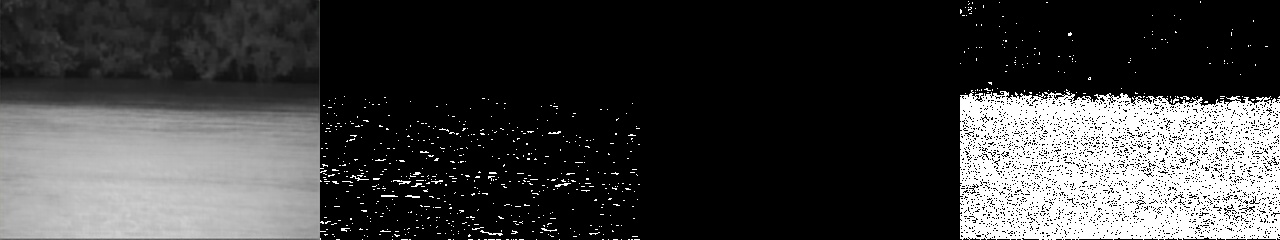}
\vspace{1pt}
\includegraphics[width=\textwidth,height=0.11\textheight]{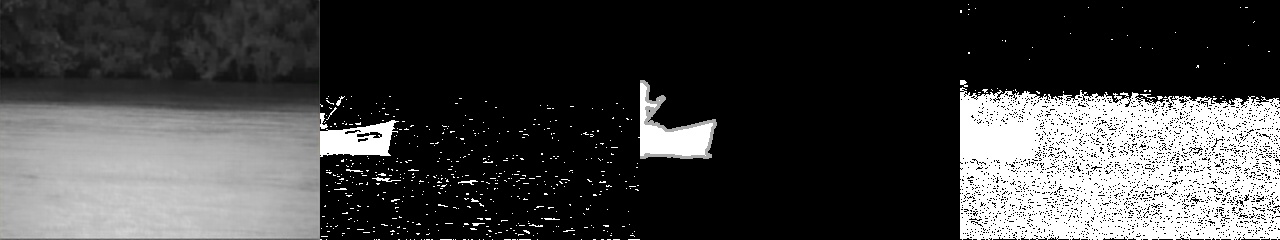}
\vspace{1pt}
\includegraphics[width=\textwidth,height=0.11\textheight]{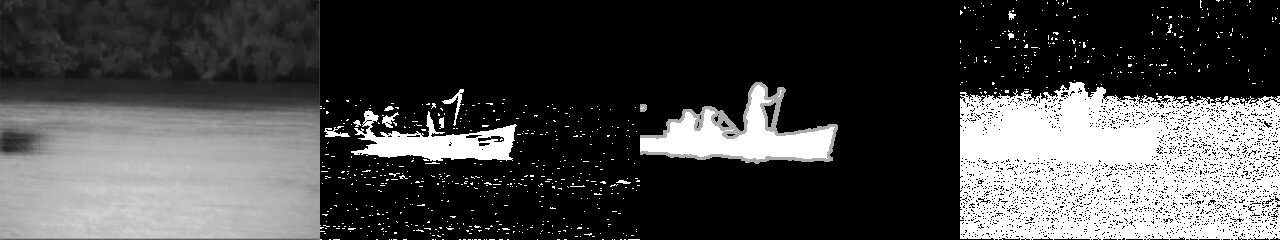}
\vspace{1pt}
\includegraphics[width=\textwidth,height=0.11\textheight]{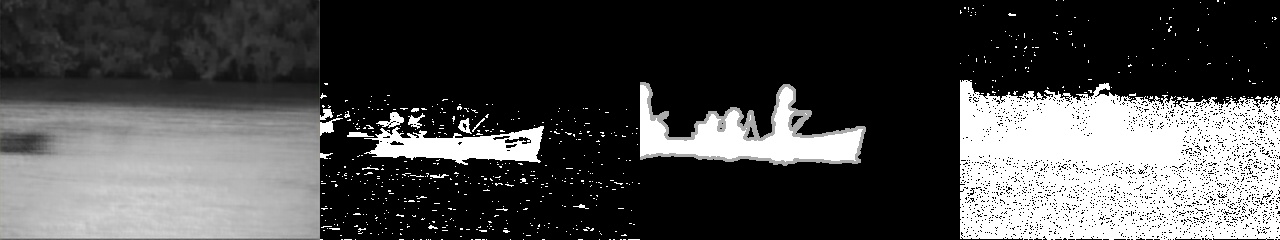}
\vspace{1pt}
\includegraphics[width=\textwidth,height=0.11\textheight]{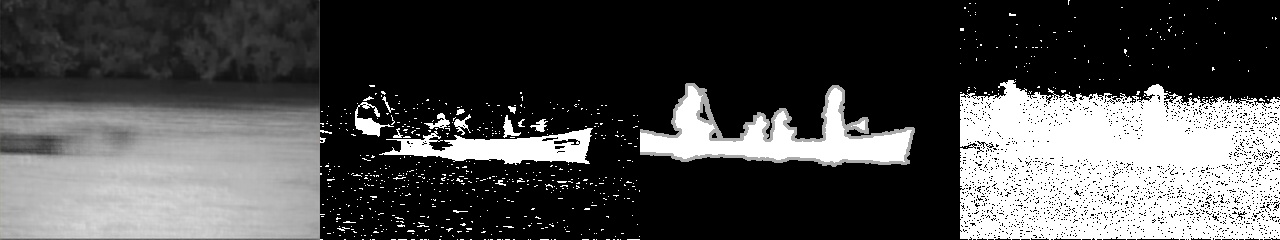}
\vspace{1pt}
\includegraphics[width=\textwidth,height=0.11\textheight]{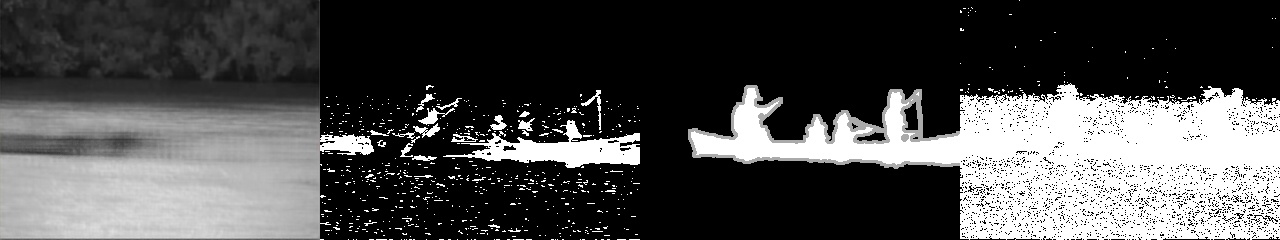}
\vspace{1pt}
\includegraphics[width=\textwidth,height=0.11\textheight]{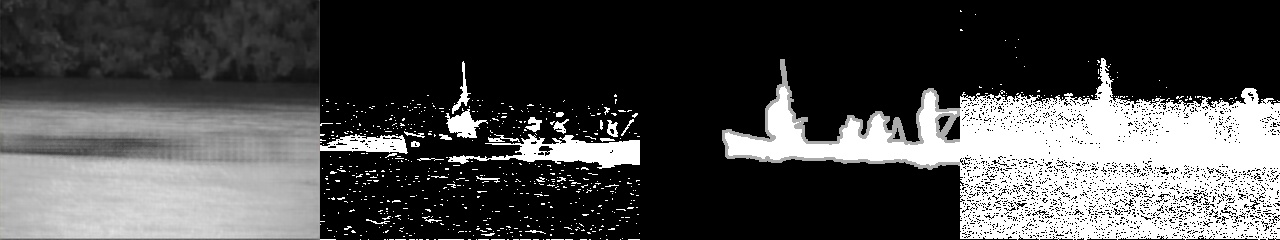}
\vspace{1pt}
\includegraphics[width=\textwidth,height=0.11\textheight]{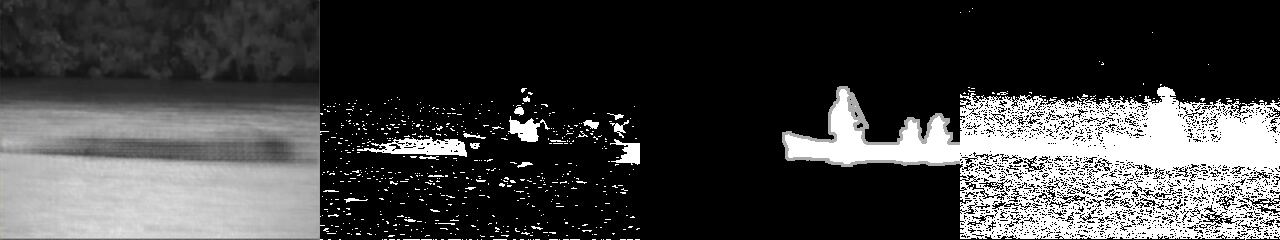}
\end{center}
\vspace{-5pt}
\caption{``Canoe'' video-sequence, cdnet~2014. \textit{Left to right}: reconstructed background outputted by the \texttt{Autoencoder}, foreground mask by the \texttt{Autoencoder}, the ground-truth, and foreground mask obtained by \texttt{OMoGMF}. This is another difficult case of non-stationary background. Notice the ``leakage'' of slowly moving object into reconstructed background. Foreground mask obtained from autoencoder output does not undergo any post-processing.}
\label{fig:canoe}
\end{figure*}

\end{document}